\documentclass[10pt,twocolumn,letterpaper]{article}

\usepackage[pagenumbers]{wacv} 

\definecolor{wacvblue}{rgb}{0.21,0.49,0.74}

\usepackage{times}
\usepackage{epsfig}
\usepackage{graphicx}
\usepackage{amsmath}
\usepackage{amssymb}
\usepackage{booktabs}
\usepackage{multirow}
\usepackage{multicol}
\usepackage{caption}
\usepackage{array}
\usepackage{colortbl}
\usepackage[table]{xcolor}
\usepackage{subcaption}
\usepackage{bm}
\usepackage{soul, color, xcolor}
\usepackage[accsupp]{axessibility}

\usepackage[pagebackref,breaklinks,colorlinks,allcolors=wacvblue]{hyperref}

\def\wacvPaperID{252} 
\def\confName{WACV}
\def\confYear{2026}

\title{CanKD: Cross-Attention-based Non-local operation \\
for Feature-based Knowledge Distillation}

\author{Shizhe Sun, Wataru Ohyama \\
Tokyo Denki University, Tokyo, Japan\\
{\tt\small yukarari77@outlook.com, w.ohyama@mail.dendai.ac.jp}
}
\begin{document}
\maketitle
\begin{abstract}
We propose Cross-Attention-based Non-local Knowledge Distillation (CanKD), a novel feature-based knowledge distillation framework that leverages cross-attention mechanisms to enhance the knowledge transfer process. Unlike traditional self-attention-based distillation methods that align teacher and student feature maps independently, CanKD enables each pixel in the student feature map to dynamically consider all pixels in the teacher feature map. This non-local knowledge transfer more thoroughly captures pixel-wise relationships, improving feature representation learning. Our method introduces only an additional loss function to achieve superior performance compared with existing attention-guided distillation methods. Extensive experiments on object detection and image segmentation tasks demonstrate that CanKD outperforms state-of-the-art feature and hybrid distillation methods. These experimental results highlight CanKD's potential as a new paradigm for attention-guided distillation in computer vision tasks. Code is available at \href{CanKD}{https://github.com/tori-hotaru/CanKD}.
\end{abstract}

\section{Introduction}
Rapid advances in deep learning have resulted in models of greater depth and width that achieve superior performance in data processing and analysis \cite{he2016deep,kirillov2023segment,liu2021swin}.
Particularly, dense prediction tasks in the field of computer vision, such as semantic segmentation \cite{zhao2017pyramid,chen2017rethinking} and object detection \cite{lin2020focal,tian2022fully}, require models to comprehend image features at the pixel level, and the inherent complexity of these features enables exceptional model performance for complex, heavy-parameter tasks. 
However, this increased model width and depth pose significant challenges to computational and memory resources. 

Knowledge distillation (KD) \cite{hinton2015distilling,buciluǎ2006model} is a viable method for addressing these challenges, enabling knowledge transfer from a high-performance, heavy-parameter teacher model to a moderate-performance, reduced-parameter student model. 
KD is broadly categorized into two approaches: one based on model output, known as logit distillation \cite{huang2022knowledge,zhao2022decoupled,ding2019adaptive}, and the other based on the intermediate feature layers of the model, known as feature distillation \cite{romero2014fitnets}. 
Instead of directly transferring the final outputs to the student model, the teacher model conveys information in logarithmic form or through intermediate layer features. 
Compared with distilling the output layer of the teacher model, distilling intermediate layers typically achieves higher accuracy, as the intermediate feature layers of the teacher model are believed to contain more information \cite{wang2021distilling}. 

\begin{figure}[t]
\centering
\includegraphics[width=\linewidth]{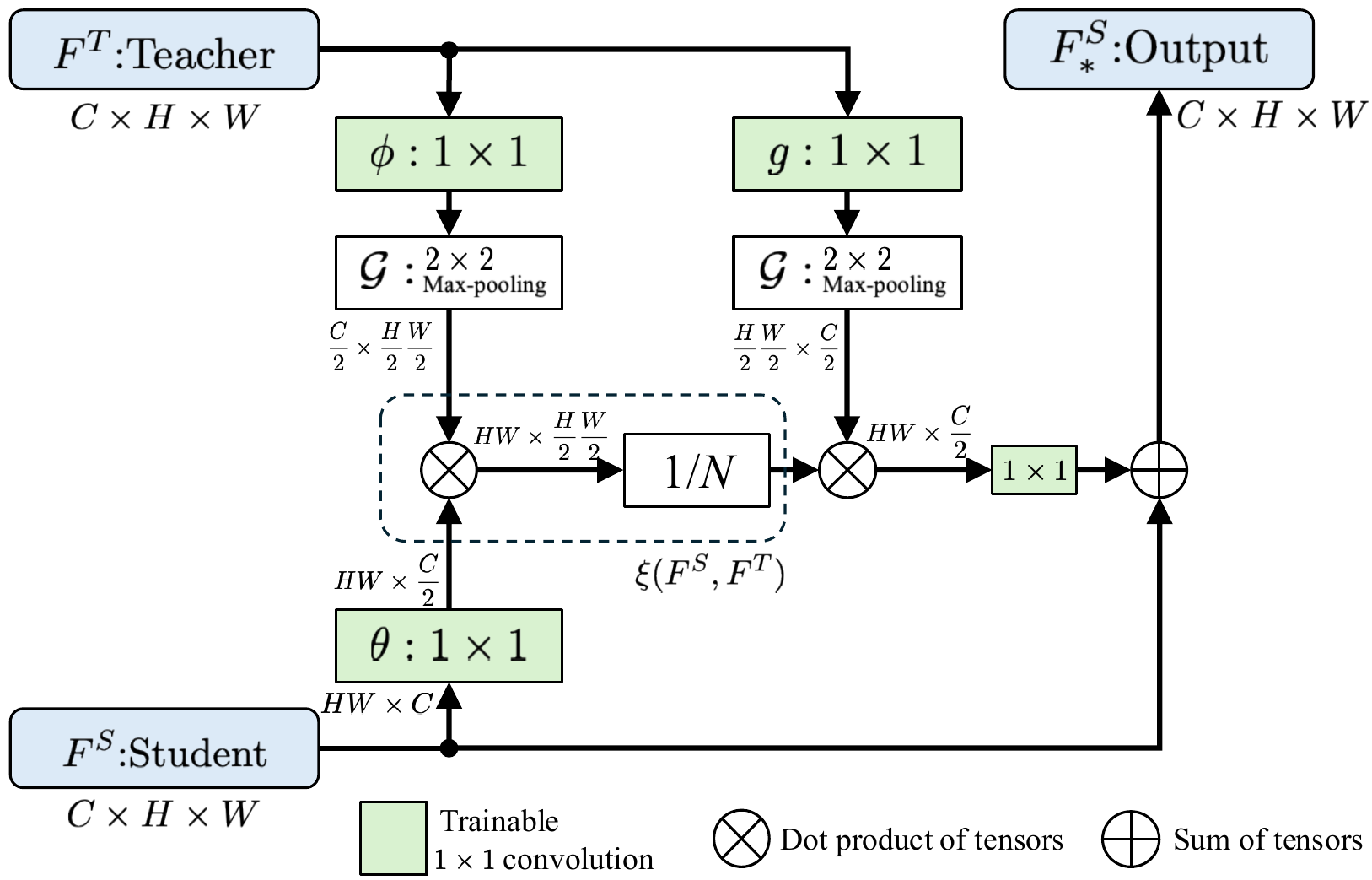}
\caption{\textbf{Proposed Cross-attention-based Non-local (Can) block.}}
\label{fig:CanBlock}
    \vspace{-5mm}
\end{figure}

Traditional knowledge distillation methods overlook the relationships between different pixels, which are beneficial for further enhancing the model's distillation performance.
To incorporate these relationships and benefit from contextual information, recent studies have employed global attention mechanisms, such as GcBlock \cite{cao2019gcnet} and Relation Networks \cite{hu2018relation}. These mechanisms utilize attention maps to establish connections between individual pixels in feature maps, potentially further enhancing the performance of the student network.
Conventional attention-based distillation models independently compute self-attention on the teacher and student feature maps before aligning them. This approach achieves consistency between pixels at corresponding positions in the student and teacher feature maps. 

Inspired by the strong performance of cross-attention in numerous dense prediction models \cite{kirillov2023segment, ravi2024sam},
we propose a hypothesis: \textbf{Whether, in a dense prediction task, cross-attention capturing relationships between each student pixel at specific locations and all teacher pixels could enhance distillation performance.}
Interestingly, in dense prediction distillation tasks, prior attention-based distillation methods have predominantly focused on self-attention, as though an inherent obstacle has impeded the use of cross-attention in this context. We posit that \textbf{the limitations of cross-attention performance may stem from the conventional attention computation schemes that have long been employed in Transformers.}

To prove our hypothesis, we propose a novel \textbf{C}ross-\textbf{A}ttention-based \textbf{N}on-local distillation method, called \textbf{CanKD}. Our method utilizes Cross-Attention Non-local (Can) blocks, as illustrated in Figure \ref{fig:CanBlock}, to enhance the student feature map and more comprehensively learn from the teacher feature map. Through cross-attention, each pixel in the student feature map can dynamically consider all pixels at various locations within the teacher feature map, thereby capturing and establishing complex pixel-level relationships between the student and teacher models. Meanwhile, we replace the conventional attention computation with a more direct method that expands the attention map values to infinity.

Compared to traditional attention-based distillation modules, such as \cite{yang2022focal,zhang2020improve}, where both the teacher and student feature maps undergo attention computations, our approach achieves superior performance with fewer parameters. \footnote{We compare CanKD with other SOTA methods in parameters and FLOPs, which are shown in Section 10 in supplementary.}
Moreover, compared with other high-performing yet complex knowledge distillation methods \cite{zhang2024freekd, huang2023knowledge}, our distillation approach introduces only an additional loss function to achieve nearly equivalent or superior performance. Conversely, in FreeKD \cite{zhang2024freekd}, the student and teacher models include L2 loss in the neck and smooth L1 loss in the head position in addition to the FreeKD loss.

We summarize our main contributions as follows:
\begin{enumerate}
    \item We present a novel Cross-Attention-based Non-local distillation method (CanKD) to enable the student model to learn from the teacher model more thoroughly, challenging the traditional paradigm in current attention-guided distillation.
    \item We evaluate the effectiveness of our method for image detection by testing multiple detector models on the COCO \cite{lin2014microsoft} dataset. Our approach outperforms state-of-the-art feature distillation methods. Meanwhile, we evaluate the performance of our method for image segmentation tasks using the Cityscapes dataset \cite{cordts2016cityscapes}, further demonstrating the general applicability of our method to image-related tasks.

    \item We evaluate the effectiveness of our method on two recent vision foundation models and one large-scale dataset. Each result exceeds the student baseline model, further confirming the universality of our approach across different model architectures and datasets of varying scales.
    
\end{enumerate}
\section{Related work}
\subsection{Knowledge distillation on dense prediction tasks}
In recent years, several studies applied knowledge distillation to dense prediction tasks. For example, MIMIC \cite{li2017mimicking}, as an early study, used L2 loss to force the intermediate layers of the student network to imitate those of the teacher network; CWD \cite{shu2021channel} processed the activation map of each channel within the teacher and student feature map to derive a soft probability map, rather than conventionally handling activation values at each spatial location; MGD \cite{yang2022masked} introduced a novel distillation approach that randomly masked the pixels of the student feature map and subsequently used simple convolution blocks to enforce the generation of the teacher's complete features; DiffKD \cite{huang2023knowledge} used a diffusion model to denoise the student feature map, thereby enabling distillation of the refined clean features and the teacher features. 

Attention-guided methods are a viable approach to enhance distillation performance in dense prediction tasks with an imbalance between foreground and background pixels. For example, FKD \cite{zhang2020improve} applied a non-local attention block to both the teacher and student feature maps. FGD \cite{yang2022focal} innovatively partitions the image into foreground and background segments and subsequently applies attention mechanisms to each segment individually. These approaches achieved state-of-the-art (SOTA) performance in dense prediction tasks.

\subsection{Cross-attention mechanism}
The advent of self-attention mechanisms \cite{vaswani2017attention} has significantly advanced the development of artificial intelligence models. Models based on self-attention mechanisms demonstrate exceptional performance in both natural language and image processing tasks \cite{devlin2018bert,brown2020language,dosovitskiy2020image,kirillov2023segment}. 
Variants of self-attention are a significant research topic.

Cross-attention was initially introduced as a variant of attention mechanisms in the decoder component of Transformer models \cite{vaswani2017attention} and progressively garnered increased popularity \cite{gheini2021cross}. 
Cross-attention effectively integrates sequences from distinct information sources by combining the queries, keys, and values of two sequences, thereby establishing dynamic associations between them. 
In computer vision, cross-attention is extensively employed in text-to-image generation tasks. 
For instance, in the Stable Diffusion model \cite{rombach2022high}, cross-attention maps various conditional modalities to the intermediate layers of the U-net architecture. 
Moreover, several models achieved highly favorable results in dense prediction and style transfer by applying cross-attention mechanisms \cite{lin2022cat,zhou2022cross}.

\section{Feature-based knowledge distillation using Cross-Attention Non-local operation}
\begin{figure*}[ht]
\centering
\begin{minipage}[b]{1.1\columnwidth}
\includegraphics[width=\linewidth]{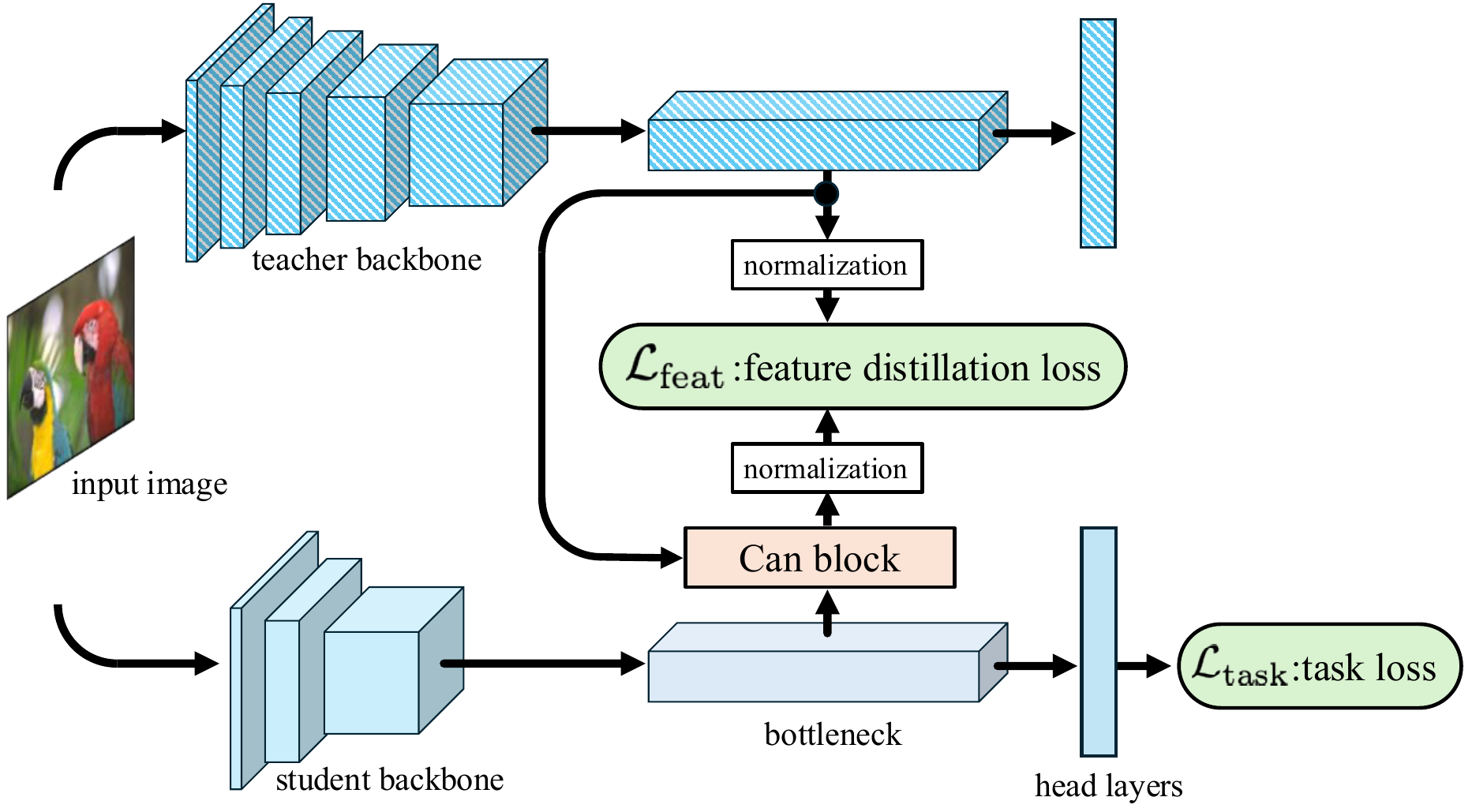}
\subcaption{\textbf{Overview of the proposed feature-based knowledge distillation framework.}}
\label{fig:ModelConf}
\end{minipage}
\hspace{0.05\columnwidth} %
\begin{minipage}[b]{0.9\columnwidth}
\includegraphics[width=0.9\columnwidth]{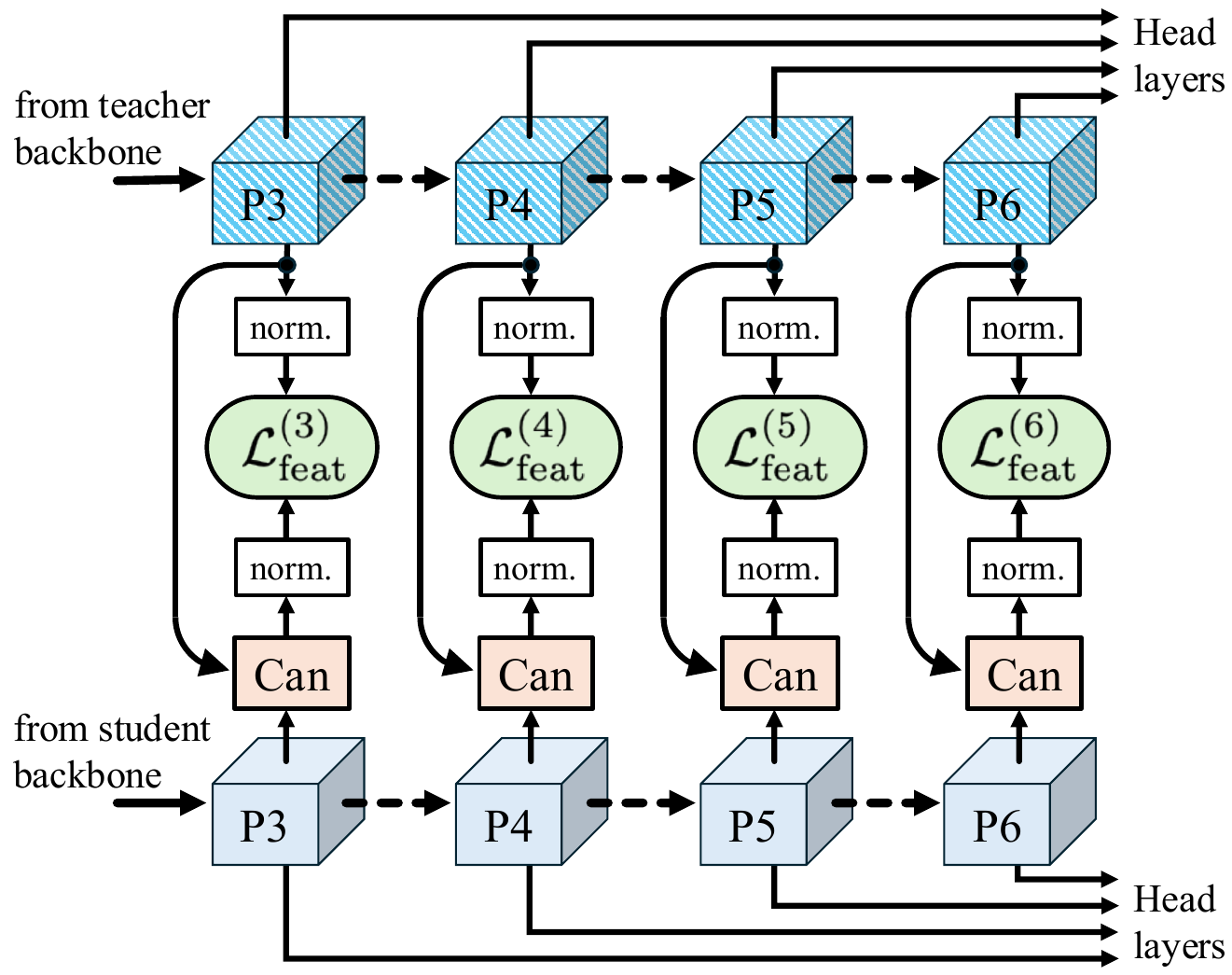}
\subcaption{\textbf{Feature-based knowledge distillation for an object detection task.} The proposed method uses the feature pyramid network (FPN) \cite{lin2017feature} for the bottleneck layers. We apply the distillation at each layer of FPN.}
\label{fig:CanKD_FPN}
\end{minipage}
\caption{\textbf{Proposed feature-based knowledge distillation using the Cross-Attention-based Non-local operation}}
\vspace{-5mm}
\end{figure*}

Feature-based knowledge distillation generally involves optimizing a loss function \(\mathcal{L}\) defined by the weighted sum of task loss \(\mathcal{L}_{\text{task}}\) and feature distillation loss \(\mathcal{L}_{\text{feat}}\).
A simple and widely used determination of \(\mathcal{L}_{\text{feat}}\) is the \(L2\)-distance between teacher and student feature maps:
\begin{equation}
\mathcal{L}_{\text{feat}}=\|\bm{F}_T-\bm{F}_S \|^2_2 ,
\label{eq:feature_distil_loss}
\end{equation}
where \(\bm{F}_T\) and \(\bm{F}_S\) denote the teacher and student features maps, respectively. 
Our proposed CanKD approach transforms the student's feature map \(\bm{F}_S\) in equation \eqref{eq:feature_distil_loss} before calculating the feature distillation loss. This enhances the guidance from the teacher model to the student model, resulting in improved feature-based knowledge distillation. Figure \ref{fig:ModelConf} illustrates the overview of CanKD. The feature map transformation using the Cross-Attention Non-local operation, which is the core of the proposed method, is denoted by the Can block on the student side of the bottleneck layers, between the backbone network and the head layers. The Can block is a specific implementation of the Can operator that introduces the operation anywhere in a neural network model. The following subsections provide a detailed description of the formulation and implementation of the Can operation.

\subsection{Cross-Attention-based Non-local operation}
The Cross-Attention Non-local operation takes two feature maps as input: an input (student) feature map \(\bm{X}\) and a reference (teacher) feature map \(\bm{Y}\), and outputs a single feature map \(\bm{Z}\). Each feature map is a sequence of spatially or temporally arranged feature vectors:
\begin{equation}
\bm{X} = \left[\bm{x}_i\right], \quad \bm Y = [\bm y_i], \quad\bm Z = \left[\bm z_i\right],
\label{eq:featuremaps_in_Can}
\end{equation}
where \(i = (1,\cdots,N)\) is a location index. \(N\) denotes the size of feature maps. For image-based tasks, \(N\) equals the product of the width \(W\) and height \(H\) of the feature map. \(\bm X, \bm Y,\) and \(\bm Z\) are of equal size.
The Can operation captures the relationship between a feature vector at a specific location in the input feature map and the feature vectors corresponding to all locations in the reference feature map. It then transforms the input feature map using the captured relationship information.

The Can operation for a feature vector at the \(i\)-th location in the input map is defined by
\begin{equation}
\bm{z}_i = \frac{1}{N}\sum_j\xi(\bm{x}_i,\bm{y}_j)g(\bm{y}_j),
\label{eq:can_operation}
\end{equation}
where \(\bm{x}_i \) and \(\bm{y}_i \) are feature vectors at the \(i\)-th location in the input (student) and reference (teacher) feature maps, respectively. \(\bm z_i\) is the output feature vector of the same size as \(\bm x\) and \(\bm y\).
\(\xi\) is a scalar function representing the relationship (similarity) between \(\bm x_i\) and \(\bm y_j\). 
\(g\) is a unary function representing a vector \(\bm y\).
This is inspired by the work in \cite{wang2018non}. However, while \cite{wang2018non} calculates self-attention within the same data source, the Can operation instead assigns feature vectors obtained from different sources to the input and reference, respectively, to compute the relationship.

Some variations exist in the definition of the \(\xi\) function. 
In this study, to facilitate implementation, we use the vector product of the embedded representations of the input and reference vectors:
\begin{equation}
\xi(\bm x_i, \bm y_j) = \theta(\bm x_i)^\top \phi(\bm y_j).
\label{eq:relation_function}
\end{equation}

\subsection{Implementation of the Can operation}
We use the Can operation for feature-based knowledge distillation in dense image-based tasks.
We implement this operation as a module, the Can block, to facilitate its incorporation into neural networks.
Figure \ref{fig:CanBlock} illustrates the structure of the Can block.
In dense image-based tasks, feature maps are originally represented as third-order tensors, denoted as \(\bm{F} \in \mathbb{R}^{H\times W\times C}\). 
We apply the Can operation by first regarding these third-order tensors as second-order tensors \(\bm{F} \in \mathbb{R}^{HW\times C}\) and then treating the student and teacher feature maps \(\bm{F}_S, \bm{F}_T\) as the input and reference feature maps \(\bm{X}, \bm{Y}\) in equation \eqref{eq:featuremaps_in_Can}, respectively. 
Here, \(H, W\), and \(C\) represent the height, width of the feature map, and the number of channels, respectively, which determine the shape of the feature maps. 

Following \cite{wang2018non}, we define the Can block as:

\begin{equation}
    \bm{F}_S^* = \bm{W}_Z\bm{Z} + \bm{F}_S,
\end{equation}
thereby combining the Can operation with a residual connection.
Each feature vector in the feature map \(\bm{Z}\) is calculated by equation \eqref{eq:can_operation}.
A residual connection performs an element-wise summation of the original student feature maps and the output feature maps of the Can operation. 
This enhances the student feature map to effectively incorporate the knowledge from various spatial locations of the teacher feature maps.\footnote{The ablation study of residual connection is shown in Section 5 in the supplementary.}

Inspired by the Transformer's decoder in \cite{vaswani2017attention}, we employ three trainable $1\times1$ convolutions in space to embed functions \(\theta, \phi\), and \(g\) in the Can operation. These convolutions perform the cross-attention mechanism to map the student feature map using $\theta$ and the teacher feature map using $\phi$ and $g$:

\begin{equation}
    \begin{aligned}
        \theta(\bm{F}_S) = \bm W_{\theta}\bm F_S,\\
        \phi(\bm F_T) = \bm W_{\phi}\bm F_T,\\
        {g}(\bm F_T) = \bm W_{g}\bm F_T,
    \end{aligned}
\end{equation}
where $\bm W_{\theta}$, $\bm W_{\phi}$, and $\bm W_{{g}}$ are \(1\times 1\) convolution weights to be trained.
As shown in equation \eqref{eq:relation_function}, we use dot product similarity as the pairwise function to calculate the affinity. This function can be easily extended to feature maps by utilizing the tensor dot product.

Additionally, we incorporate a down-sampling technique in our block to reduce pairwise computation. We apply a max pooling layer to each of the two embeddings, $\phi(\bm{F}_T)$ and ${g}(\bm{F}_T)$, derived from the teacher feature maps. This technique can reduce pairwise computation.\footnote{The ablation study of max pooling is available in Section 4 in the supplementary.}

\subsection{Overall loss}
We posit that a discrepancy exists between the distributions of the enhanced student and teacher feature maps. This discrepancy can, to some extent, impact the model's distillation performance. To mitigate this discrepancy, we incorporate instance normalization to standardize both feature maps before computing the L2 loss. Thus, our feature distillation loss function is:
\begin{equation}
    \mathcal{L}_\text{feat}=\|\Omega(\bm{F}_T)-\Omega(\bm{F}_S^*)\|^2_2,
\end{equation}
where $\Omega()$ is the normalization process.

We introduce a hyperparameter $\mu$ to balance the loss of feature distillation and the loss of task. 
The overall training loss to optimize is:
\begin{equation}
    \label{eq:total}
    \mathcal{L}_\text{student} = \mathcal{L}_\text{task} + \mu\mathcal{L}_\text{feat}.
\end{equation}
Notably, our distillation loss is primarily applied to intermediate feature maps, such as those used in Feature Pyramid Networks (FPN) \cite{lin2017feature}, making it applicable to various architectures. Furthermore, our distillation method merely introduces an additional loss term to the existing loss function, enabling its seamless integration with other distillation approaches.

\section{Experiments}
We applied the proposed method to two types of dense tasks, namely object detection and semantic segmentation, to evaluate its effectiveness. We followed the standard experimental procedures commonly used in previous studies for both tasks and compared our performance with that of state-of-the-art approaches.\footnote{We also test classification task with ImageNet-1k, which are shown in Section 9 in supplementary.}
\subsection{Object detection}
\subsubsection{Details}
We verify our method on the MSCOCO detection dataset \cite{lin2014microsoft}, which contains 80 object classes. We train our student models on COCO train2017 and test on COCO val2017. We use average precision (AP) to evaluate performance.

As in \cite{huang2023knowledge,zhang2020improve,yang2022focal}, we experiment using baseline and enhanced settings. In the baseline setting, we test two-stage detector Faster-RCNN \cite{ren2015faster}, one-stage detector RetinaNet \cite{lin2017focal}, and anchor-free detector FCOS \cite{tian2019fcosfullyconvolutionalonestage}. In the teacher network and student network, the backbone is ResNet-101 (R101) \cite{he2016deep} and ResNet-50 (R50), respectively. In the enhanced setting, we replace ResNet-101 with ResNeXt101 (X101) \cite{xie2017aggregated} in the teacher backbone. We test two-stage detector Cascade Mask RCNN \cite{cai2018cascade}, one-stage detector RetinaNet \cite{lin2017focal}, and anchor-free detector RepPoints \cite{yang2019reppoints}. These methods conduct the distillation in the model's neck module. As shown in Figure \ref{fig:CanKD_FPN}, we distill the output of every layer within the feature pyramid located in the neck (from P3 to P6).

Following the official strategies in MMDetection \cite{mmdetection}, we use stochastic gradient descent (SGD) with a learning rate of 0.005 and weight decay of 1e-4 with $2\times$ schedule in MMRazor \cite{2021mmrazor}.\footnote{The further details of training strategies are shown in Section 1 in the supplementary.} The loss weights $\mu$ in equation \eqref{eq:total} are all set to five, as determined by an ablation study. Furthermore, to verify that our method maintains suitable performance with a heterogeneous teacher, we follow the PKD \cite{cao2022pkd} training strategy, using FCOS-X101 as the teacher and RetinaNet-R50 as the student to conduct additional experiments.

\begin{table}[tb]
    \centering
        \caption{\textbf{Object detection on the COCO validation dataset under baseline setting.} T:Teacher. S:Student. $\dagger$ denotes \textit{inheriting strategy} \cite{kang2021instance, lidetkds, zhang2024freekd}. F:Feature distillation. L:logit distillation. $\ddagger$ denotes introducing the space search algorithm\cite{li2023automated,lidetkds}. We train FCOS\cite{tian2019fcosfullyconvolutionalonestage} with techniques similar to those in FGD \cite{yang2022focal}.}
    \label{1}
    \resizebox{\linewidth}{!}{
        \begin{tabular}{lcccccccc}
        \toprule
        \multicolumn{2}{l}{Method} & Distillation & $AP$ & $AP_{50}$ & $AP_{75}$ & $AP_S$ & $AP_M$ & $AP_L$ \\
        \midrule
        \midrule
        \multicolumn{9}{c}{\it{One-stage detectors}} \\
        \midrule
        \multicolumn{2}{l|}{T: RetinaNet-R101} & \textit{N/A} & 38.9 & 58.0 & 41.5 & 21.0 & 42.8 & 52.4 \\
        \multicolumn{2}{l|}{S: RetinaNet-R50}  & \textit{N/A} & 37.4 & 56.7 & 39.6 & 20.0 & 40.7 & 49.7 \\
        \midrule
        \multicolumn{2}{l|}{FGD\cite{yang2022focal} }               & F & 39.6(2.2$\uparrow$) & - & - & \bf{22.9} & 43.7 & 53.6 \\
        \multicolumn{2}{l|}{Auto-KD\cite{li2023automated} }               & F+L$\ddagger$ & 39.5(2.1$\uparrow$) & 58.5 & 42.4 & 22.8 & 43.5 & 53.3 \\
        \multicolumn{2}{l|}{DiffKD\cite{huang2023knowledge} }            & F+L & 39.7(2.3$\uparrow$) & 58.6 & 42.1 & 21.6 & 43.8 & 53.3 \\
        \midrule
        \rowcolor{gray!20}
        \multicolumn{2}{l|}{CanKD}            & F & 39.7(2.3$\uparrow$) & 58.6 & 42.5 & 21.8 & 43.8 & 53.6 \\
        \rowcolor{gray!20}
        \multicolumn{2}{l|}{CanKD$\dagger$}    & F & \textcolor{blue}{\bf{39.8(2.4$\uparrow$)}} & 58.8 & \bf{42.6} & 21.6 & \bf{44.0} & \bf{53.7} \\
        \midrule
        \multicolumn{9}{c}{\it{Two-stage detectors}} \\
        \midrule
        \multicolumn{2}{l|}{T: FasterRCNN-R101} & \textit{N/A}& 39.8 & 60.1 & 43.3 & 22.5 & 43.6 & 52.8 \\
        \multicolumn{2}{l|}{S: FasterRCNN-R50}  & \textit{N/A}& 38.4 & 59.0 & 42.0 & 21.5 & 42.1 & 50.3 \\
        \midrule
        \multicolumn{2}{l|}{FGD\cite{yang2022focal} }               & F & 40.4(2.0$\uparrow$) & - & - & 22.8 & 44.5 & 53.5 \\
        \multicolumn{2}{l|}{PKD\cite{cao2022pkd} }               & F & 40.5(2.1$\uparrow$) & \bf{60.9} & \bf{44.4} & 22.6 & 44.8 & 53.1 \\
        \multicolumn{2}{l|}{Auto-KD\cite{li2023automated} }             & F+L$\ddagger$  & 40.1(1.7$\uparrow$) & 60.6 & 43.7 & 22.7 & 44.0 & 52.8 \\
        \multicolumn{2}{l|}{DiffKD\cite{huang2023knowledge} }          & F+L  & 40.6(2.2$\uparrow$) & \bf{60.9} & 43.9 & 23.0 & 44.5 & \bf{54.0} \\
        \multicolumn{2}{l|}{DetKDS$\dagger$\cite{lidetkds} }           & F$\ddagger$  & 40.6(2.2$\uparrow$) & 60.8 & 44.1 & 22.9 & 44.8 & 53.5 \\
        \midrule
        \rowcolor{gray!20}
        \multicolumn{2}{l|}{CanKD}             & F & 40.1(1.7$\uparrow$) & 60.4 & 43.7 & 22.1 & 44.3 & 52.8 \\
        \rowcolor{gray!20}
        \multicolumn{2}{l|}{CanKD$\dagger$}    & F & \textcolor{blue}{\bf{40.7(2.3$\uparrow$)}} & 60.8 & \bf{44.4} & \bf{23.3} & \bf{44.9} & 53.6 \\
        \midrule
        \multicolumn{9}{c}{\it{Anchor-free detectors}} \\
        \midrule
        \multicolumn{2}{l|}{T: FCOS-R101} & \textit{N/A} & 40.8 & 60.0 & 44.0 & 24.2 & 44.3 & 52.4 \\
        \multicolumn{2}{l|}{S: FCOS-R50} & \textit{N/A} & 38.5 & 57.7 & 41.0 & 21.9 & 42.8 & 48.6 \\
        \midrule
        \multicolumn{2}{l|}{FGD\cite{yang2022focal} }              & F & 42.1(3.6$\uparrow$) & - & - & 27.0 & 46.0 & 54.6 \\
        \multicolumn{2}{l|}{Auto-KD\cite{li2023automated} }              & F+L$\ddagger$ & 42.0(3.5$\uparrow$) & 60.4 & 45.5 & 25.9 & 45.8 & 54.4 \\
        \multicolumn{2}{l|}{DiffKD\cite{huang2023knowledge} }          & F+L  & \textcolor{blue}{\bf{42.4(3.9$\uparrow$)}} & \bf{61.0} & 45.8 & 26.6 & 44.5 & 54.0 \\
        \multicolumn{2}{l|}{PCKA+mimic\cite{ijcai2024p628} }          & F+L  & 40.7(3.3$\uparrow$) & 60.5 & 43.1 & 23.4 & 44.8 & 53.1 \\
        \multicolumn{2}{l|}{DetKDS$\dagger$\cite{lidetkds} }           & F$\ddagger$ & 42.3(3.8$\uparrow$) & 60.8 & 45.6 & 26.1 & 46.2 & 54.7 \\
        \midrule
        \rowcolor{gray!20}
        \multicolumn{2}{l|}{CanKD}            & F & \textcolor{blue}{\bf{42.4(3.9$\uparrow$)}} & 60.8 & \bf{45.9} & 26.1 & \bf{46.5} & 55.0 \\
        \rowcolor{gray!20}
        \multicolumn{2}{l|}{CanKD$\dagger$}   & F & \textcolor{blue}{\bf{42.4(3.9$\uparrow$)}} & 60.8 & 45.8 & \bf{26.8} & 46.1 & \bf{55.2} \\
        \bottomrule
        \end{tabular}
    }
\end{table}

\begin{table}[tb]
    \centering
    \caption{\textbf{Object detection on the COCO validation dataset under enhanced setting.} T:Teacher. S:Student. \textit{CM RCNN}: Cascade Mask RCNN. $\dagger$ denotes \textit{inheriting strategy} \cite{kang2021instance, lidetkds, zhang2024freekd}. F:Feature distillation. L:logit distillation. $\ddagger$ denotes introducing the space search algorithm\cite{li2023kd,lidetkds}.}
    \label{2}    
    \resizebox{\linewidth}{!}{
        \begin{tabular}{lcccccccc}
        \toprule
        \multicolumn{2}{l}{Method} & Distillation & $AP$ & $AP_{50}$ & $AP_{75}$ & $AP_S$ & $AP_M$ & $AP_L$ \\
        \midrule
        \midrule
        \multicolumn{9}{c}{\it{One-stage detectors}} \\
        \midrule
        \multicolumn{2}{l|}{T: RetinaNet-X101} & \textit{N/A} & 41.0 & 60.9 & 44.0 & 23.9 & 45.2 & 54.0 \\
        \multicolumn{2}{l|}{S: RetinaNet-R50}  & \textit{N/A} & 37.4 & 56.7 & 39.6 & 20.0 & 40.7 & 49.7 \\
        \midrule
        \multicolumn{2}{l|}{FKD\cite{zhang2020improve} }              & F & 39.6(2.2$\uparrow$) & 58.8 & 42.1 & 22.7 & 43.3 & 52.5 \\
        \multicolumn{2}{l|}{FGD\cite{yang2022focal} }            & F& 40.4(3.0$\uparrow$) & - & - & 23.4 & 44.7 & 54.1 \\
        \multicolumn{2}{l|}{KD-Zero\cite{li2023kd} }         &  F+L$\ddagger$ & 40.9(3.5$\uparrow$) & 60.4 & 43.5 & 23.2 & 45.2 & 54.8 \\
        \multicolumn{2}{l|}{DiffKD\cite{huang2023knowledge} }        &   F+L & 40.7(3.3$\uparrow$) & 60.0 & 43.2 & 22.2 & 45.0 & 55.2 \\
        \multicolumn{2}{l|}{PCKA+mimic\cite{ijcai2024p628} }          &F+L  & 40.7(3.3$\uparrow$) & 60.4 & 43.4 & 23.9 & 44.7 & 55.1 \\
        \multicolumn{2}{l|}{DetKDS$\dagger$\cite{lidetkds} }           & F$\ddagger$ & 41.0(3.6$\uparrow$) & 60.2 & 43.6 & 23.0 & 45.2 & 55.0 \\
        \multicolumn{2}{l|}{FreeKD$\dagger$\cite{zhang2024freekd} }         &  F+L & 41.0(3.6$\uparrow$) & - & - & 22.3 & 45.1 & \bf{55.7} \\
        \midrule
        \rowcolor{gray!20}
        \multicolumn{2}{l|}{CanKD}            & F & 41.0(3.6$\uparrow$) & \bf{60.6} & \bf{43.8} & \bf{24.3} & \bf{45.9} & 54.4 \\
        \rowcolor{gray!20}
        \multicolumn{2}{l|}{CanKD$\dagger$}   & F & \textcolor{blue}{\bf{41.1(3.7$\uparrow$)}} & \bf{60.6} & \bf{43.8} & 23.6 & 45.4 & 55.3 \\
        \midrule
        \multicolumn{9}{c}{\it{Two-stage detectors}} \\
        \midrule
        \multicolumn{2}{l|}{T: CM RCNN-X101} & \textit{N/A} & 45.6 & 64.1 & 49.7 & 26.2 & 49.6 & 60.0 \\
        \multicolumn{2}{l|}{S: FasterRCNN-R50} & \textit{N/A} & 38.4 & 59.0 & 42.0 & 21.5 & 42.1 & 50.3 \\
        \midrule
        \multicolumn{2}{l|}{FKD\cite{zhang2020improve} }             & F  & 41.5(3.1$\uparrow$) & 62.2 & 45.1 & 23.5 & 45.0 & 55.3 \\
        \multicolumn{2}{l|}{FGD\cite{yang2022focal} }             & F  & 42.0(3.6$\uparrow$) & - & - & 23.7 & 46.4 & 55.5 \\
        \multicolumn{2}{l|}{KD-Zero\cite{li2023kd} }          & F+L$\ddagger$ & 41.9(3.5$\uparrow$) & 62.7 & 45.5 & 23.6 & 45.6 & 55.6 \\
        \multicolumn{2}{l|}{DiffKD\cite{huang2023knowledge} }          & F+L & 42.2(3.8$\uparrow$) & 62.8 & 46.0 & 24.2 & 46.6 & 55.3 \\
        \multicolumn{2}{l|}{PCKA+mimic\cite{ijcai2024p628} }         & F+L  & 42.4(4.0$\uparrow$) & \bf{63.3} & 46.1 & 24.3 & 46.7 & 56.1 \\
        \multicolumn{2}{l|}{DetKDS$\dagger$\cite{lidetkds} }          & F$\ddagger$ & 42.3(3.9$\uparrow$) & 62.8 & 46.2 & \bf{24.8} & 46.1 & 56.0 \\
        \multicolumn{2}{l|}{FreeKD$\dagger$\cite{zhang2024freekd} }          & F+L & 42.4(4.0$\uparrow$) & - & - & 24.1 & 46.7 & 55.9 \\
        \midrule
        \rowcolor{gray!20}
        \multicolumn{2}{l|}{CanKD}            & F & 42.0(3.6$\uparrow$) & 62.4 & 45.6 & 23.8 & 46.4 & 55.1 \\
        \rowcolor{gray!20}
        \multicolumn{2}{l|}{CanKD$\dagger$}   & F & \textcolor{blue}{\bf{42.5(4.1$\uparrow$)}} & 62.9 & \bf{46.5} & 23.9 & \bf{47.0} & \bf{56.7} \\
        \midrule
        \multicolumn{9}{c}{\it{Anchor-free detectors}} \\
        \midrule
        \multicolumn{2}{l|}{T: RepPoints-X101} & \textit{N/A} & 44.2 & 65.5 & 47.8 & 26.2 & 48.4 & 58.5 \\
        \multicolumn{2}{l|}{S: RepPoints-R50} & \textit{N/A} & 38.6 & 59.6 & 41.6 & 22.5 & 42.2 & 50.4 \\
        \midrule
        \multicolumn{2}{l|}{FKD\cite{zhang2020improve} }              & F & 40.6(2.0$\uparrow$) & 61.7 & 43.8 & 23.4 & 44.6 & 53.0 \\
        \multicolumn{2}{l|}{FGD\cite{yang2022focal} }              & F & 41.3(2.7$\uparrow$) & - & - & \bf{24.5} & 45.2 & 54.0 \\
        \multicolumn{2}{l|}{DiffKD\cite{huang2023knowledge} }          & F+L  & 41.7(3.1$\uparrow$) & 62.6 & 44.9 & 23.6 & 45.4 & 55.9 \\
        \multicolumn{2}{l|}{DetKDS$\dagger$\cite{lidetkds} }          & F$\ddagger$  & 42.3(3.7$\uparrow$) & \bf{63.1} & \bf{45.8} & 24.1 & 46.4 & 55.9 \\
        \multicolumn{2}{l|}{FreeKD$\dagger$\cite{zhang2024freekd}}          & F+L  & 42.4(3.8$\uparrow$) & - & - & 24.3 & 46.4 & \bf{56.6} \\
        \midrule
        \rowcolor{gray!20}
        \multicolumn{2}{l|}{CanKD}            & F & 42.4(3.8$\uparrow$) & 62.9 & 45.6 & 24.1 & 46.5 & 56.4 \\
        \rowcolor{gray!20}
        \multicolumn{2}{l|}{CanKD$\dagger$}  & F  & \textcolor{blue}{\bf{42.6(4.0$\uparrow$)}} & \bf{63.1} & 45.6 & 24.1 & \bf{47.0} & 56.4 \\
        \bottomrule
        \end{tabular}
    }
\end{table}

\begin{table}[tb]
    \centering
    \caption{\textbf{Object detection on the COCO validation dataset with heterogeneous teacher.} T:Teacher. S:Student.}
    \label{table}    
    \resizebox{\linewidth}{!}{
        \begin{tabular}{lccccccc}
        \toprule
        \multicolumn{2}{l}{Method} & $AP$ & $AP_{50}$ & $AP_{75}$ & $AP_S$ & $AP_M$ & $AP_L$ \\
        \midrule
        \midrule
        \multicolumn{2}{l|}{T: FCOS-X101} & 42.7 & 62.5 & 45.7 & 26.0 & 46.5 & 54.7 \\
        \multicolumn{2}{l|}{S: RetinaNet-R50}  & 37.4 & 56.7 & 39.6 & 20.0 & 40.7 & 49.7 \\
        \midrule
        \multicolumn{2}{l|}{PKD\cite{cao2022pkd} }               & 40.3(2.9$\uparrow$) & 59.6 & 43.0 & 22.2 & 44.9 & 53.7 \\
        \multicolumn{2}{l|}{DetKDS\cite{lidetkds} }            & 40.4(3.0$\uparrow$) & 59.8 & 43.0 & 22.8 & 44.3 & 53.8 \\
        \midrule
        \rowcolor{gray!20}
        \multicolumn{2}{l|}{CanKD}             & \textcolor{blue}{\bf{40.5(3.1$\uparrow$)}} & \bf{59.9} & \bf{43.4} & \bf{23.2} & \bf{45.1} & \bf{54.2} \\
        \bottomrule
        \end{tabular}
    }
    \vspace{-3mm}
\end{table}

\subsubsection{Experiment results}
We compared existing KD methods in Table \ref{1} using the baseline setting. 
Our CanKD approach significantly improves the performance of the student model across multiple frameworks, surpassing other state-of-the-art feature distillation methods and even exceeding the performance of specific state-of-the-art hybrid distillation approaches. 
For instance, CanKD using FasterRCNN \cite{ren2015faster} with a ResNet-R50 \cite{he2016deep} backbone attains a 2.3-point AP improvement over the baseline on COCO. 

We further investigate the effectiveness of CanKD using an enhanced teacher model. 
Our results are summarized in Table \ref{2}. We observe that CanKD demonstrates even greater superiority when applied with more powerful teacher models. Notably, CanKD achieves a 4.0-point AP improvement over the baseline when using the RepPoints-R50 \cite{yang2019reppoints} student model, surpassing the AP of the following best method, FreeKD, by 0.2 points. With the CanKD method, the RetinaNet-R50 \cite{lin2017focal} student model even surpasses the RetinaNet-X101 \cite{lin2017focal} teacher model on the COCO dataset. 

Under heterogeneous teacher conditions, our model also demonstrates excellent performance, achieving a 3.1-point AP improvement over the baseline and surpassing the AP of the top current state-of-the-art model DetKDS by 0.1 points, as shown in Table \ref{table}.\footnote{The model's confusion matrix and changes in training process are shown in Section 6 and 7 in the supplementary.}

\subsection{Semantic segmentation}
\subsubsection{Details}
We verify our method on the Cityscapes dataset \cite{cordts2016cityscapes}, which comprises 5,000 high-quality images (2,975 training images and 500 validation images). In this task, we evaluate performance using the mean Intersection-over-Union (mIoU) metric.

We experiment with two segmentation frameworks, PSPNet \cite{zhao2017pyramid} and DeeplabV3 \cite{chen2018encoder}. Our teacher backbone is ResNet101. Our student backbone uses two frameworks: ResNet18 and MobileNetV2 \cite{sandler2018mobilenetv2}. The crop size of all teacher and student models is $512\times 1024$. The loss weights, $\mu$, of CanKD in equation \eqref{8} are set to ten, as determined by an ablation study. In other configurations, the loss weights $\mu$ are all set to five following the configuration of the detection task. The training strategies of CWD \cite{shu2021channel} and MGD \cite{yang2022masked} follow the original CWD and MGD papers. The distillation position of CanKD and MGD are the model's backbone output, while the distillation position of CWD is the logit output. To address the channel mismatch between teacher and student feature maps, we follow prior work \cite{yang2022focal,yang2022masked,zhang2024freekd}, introducing a 1×1 convolution to align the channel dimensions of the student feature maps with those of the teacher. (We adopt the same strategy for other tasks as well.) Following the settings in MMSegmentation \cite{mmdetection}, we use SGD with a learning rate of 0.01, weight decay of 5e-4, and 80K schedule in the MMRazor \cite{2021mmrazor} framework for all methods.

\begin{table}[tb]
    \centering
    \caption{\textbf{Semantic segmentation results on the Cityscapes validation dataset.} FLOPs is measured based on an input image size of $1024\times2048$. \textit{* indicates results reproduced for this paper.}}
    \resizebox{\linewidth}{!}{
        \begin{tabular}{lccc|c|cc}
        \toprule
        \multicolumn{2}{l}{Method} & FLOPs(G) & Params(M) & mIoU(\%) & aAcc(\%) & mAcc(\%)\\
        \midrule
        \midrule
        \multicolumn{2}{l|}{T: PSPNet-R101} & 2047.11 & 65.49 & 79.74 & 96.33 & 86.56 \\
        \midrule
        \multicolumn{2}{l|}{S: PSPNet-R18}  & \multirow{4}{*}{433.86} & \multirow{4}{*}{12.63} & 74.87 & 95.56 & 82.07 \\
        \multicolumn{2}{l|}{CWD\cite{shu2021channel} }            & & & 75.54 & 95.75 & 82.11 \\
        \multicolumn{2}{l|}{MGD*\cite{yang2022masked} }            & & & 75.84 & \bf{95.68} & 83.07 \\
        \rowcolor{gray!20}
        \multicolumn{2}{l|}{CanKD}         & & & \textcolor{blue}{\bf{76.24}} & 95.67 & \bf{85.49} \\
        \midrule
        \multicolumn{2}{l|}{S: DeepLabV3-R18}  & \multirow{4}{*}{479.44} & \multirow{4}{*}{13.83} & 76.70 & 95.89 & 84.29 \\
        \multicolumn{2}{l|}{CWD*\cite{shu2021channel} }               & & & 77.19 & 95.93 & 85.35 \\
        \multicolumn{2}{l|}{MGD*\cite{yang2022masked} }               & & & 77.28 & 95.90 & 85.11 \\
        \rowcolor{gray!20}
        \multicolumn{2}{l|}{CanKD}         & & & \textcolor{blue}{\bf{77.34}} & \bf{96.04} & \bf{85.49} \\
        \midrule
        \midrule
        \multicolumn{2}{l|}{T: DeepLabV3-R101} & 2777.00 & 84.63 & 80.31 & 96.44 & 87.16 \\
        \midrule
        \multicolumn{2}{l|}{S: DeepLabV3-R18}  & \multirow{3}{*}{479.44} & \multirow{3}{*}{13.83} & 76.70 & 95.89 & 84.29 \\
        \multicolumn{2}{l|}{MGD*\cite{yang2022masked} }               & & & 76.95 & 95.90 & 84.53 \\
        \rowcolor{gray!20}
        \multicolumn{2}{l|}{CanKD}    & & & \textcolor{blue}{\bf{77.12}} & \bf{96.01} & \bf{85.47} \\
        \midrule
        \multicolumn{2}{l|}{S: DeepLabV3-MV2}  & \multirow{2}{*}{598.56} & \multirow{2}{*}{18.34} & 73.84 & 95.28 & 83.72 \\
        \multicolumn{2}{l|}{CanKD}         & & & \textcolor{blue}{\bf{75.62}} & \bf{95.94} & \bf{84.81} \\
        \bottomrule
        \end{tabular}
    }
    \label{3}
    \vspace{-5mm}
\end{table}

\subsubsection{Experiment results}
Our semantic segmentation results are recorded in Table \ref{3}. Compared with previous state-of-the-art models, CanKD demonstrates outstanding performance for high-resolution images. In homogeneous settings, CanKD achieves a 0.42-point mIoU improvement over the baseline using DeepLabV3-R18, surpassing MGD\cite{yang2022masked} by 0.17 mIoU points. Furthermore, we observe that CanKD delivers excellent performance when the teacher and student models are heterogeneous. For instance, CanKD improves the mIoU by 0.64 points over the baseline when using the DeepLabV3-R18\cite{chen2018encoder} student model with the PSPNet-R101\cite{zhao2017pyramid} teacher model.

\subsection{Recent vision foundation models and large-scale dataset}
To demonstrate the generality of our approach, we verify CanKD on two popular DETR-based models (Dino \cite{zhang2022dino} and Grounding-Dino \cite{liu2024grounding}) and a large-scale dataset (Object365v2 \cite{Shao2019object365}) with various student backbone (R50, Swin-b\cite{liu2021swin}) in Table \ref{new}. From the results, our CanKD achieves improvements of +0.7 ,+0.2 and +0.9 over the student baseline on Dino and Grounding-Dino models. Furthermore, on ultra-large Object365v2 dataset, our CanKD outperforms the baseline by +1.7, further demonstrating the generality and advancement of our method.
\begin{table}[tb]
    \centering
    \caption{\textbf{Object detection on recent vision foundation models and large-scale dataset.} T:Teacher. S:Student.}
    \label{new}    
    \resizebox{\linewidth}{!}{
        \begin{tabular}{lccccccc}
        \toprule
        \multicolumn{2}{l}{Method} & $AP$ & $AP_{50}$ & $AP_{75}$ & $AP_S$ & $AP_M$ & $AP_L$ \\
        \midrule
        \midrule
        \multicolumn{2}{l|}{T: Dino-R101} & 50.4 & 68.0 & 54.8 & 32.9 & 53.9 & 65.2 \\
        \multicolumn{2}{l|}{S: Dino-R50}  & 49.0 & 66.4 & 53.3 & 31.4 & 52.2 & 64.0 \\
        \midrule
        \multicolumn{2}{l|}{CWD\cite{shu2021channel}}             & 49.5(0.5$\uparrow$) & 67.0 & \textbf{54.2} & 31.5 & 52.7 & \textbf{64.5} \\
        \multicolumn{2}{l|}{PKD\cite{cao2022pkd}}             & 49.2(0.2$\uparrow$) & 66.7 & 53.8 & 32.3 & 52.4 & 63.8 \\
        \multicolumn{2}{l|}{QPD\cite{li2024distilling}}             & 49.5(0.5$\uparrow$) & 67.2 & 53.9 & - & - & - \\
        \multicolumn{2}{l|}{QRD\cite{li2024distilling}}             & 49.5(0.5$\uparrow$) & \textbf{67.5} & 54.1 & - & - & - \\
        \midrule
        \rowcolor{gray!20}
        \multicolumn{2}{l|}{CanKD}             & \textcolor{blue}{\bf{49.7(0.7$\uparrow$)}} & 67.1 & \textbf{54.2} & \textbf{32.5} & \textbf{52.8} & 64.4 \\
        \midrule
        \multicolumn{2}{l|}{T: Dino-Swin-L} & 58.4 & 77.1 & 64.2 & 41.5 & 62.2 & 73.5 \\
        \multicolumn{2}{l|}{S: Dino-Swin-B} & 56.3 & 74.7 & 61.8 & 40.0 & 59.8 & 73.0 \\
        \midrule
        \multicolumn{2}{l|}{CWD\cite{shu2021channel}}             & 54.7(1.6$\downarrow$) & 72.9 & 59.9 & 37.6 & 57.7 & 72.4 \\
        \midrule
        \rowcolor{gray!20}
        \multicolumn{2}{l|}{CanKD}             & \textcolor{blue}{\bf{56.5(0.2$\uparrow$)}} & \textbf{74.8} & \textbf{61.9} & \textbf{40.4} & \textbf{60.0} & 72.6 \\
        \midrule
        \midrule
        \multicolumn{2}{l|}{T: GroundingDino-R101} & 50.0 & 67.4 & 54.6 & 32.3 & 54.0 & 64.5 \\
        \multicolumn{2}{l|}{S: GroundingDino-R50}  & 48.9 & 66.4 & 53.3 & 30.7 & 52.2 & 64.2 \\
        \midrule
        \multicolumn{2}{l|}{CWD\cite{shu2021channel}}             & 49.5(0.6$\uparrow$) & 66.7 & \textbf{54.1} & \textbf{32.3} & 53.0 & 64.1 \\
        \multicolumn{2}{l|}{PKD\cite{cao2022pkd}}             & 49.4(0.5$\uparrow$) & 66.8 & 53.9 & 31.7 & 52.8 & 64.0 \\
        \midrule
        \rowcolor{gray!20}
        \multicolumn{2}{l|}{CanKD}             & \textcolor{blue}{\bf{49.8(0.9$\uparrow$)}} & \textbf{66.9} & \textbf{54.1} & 31.5 & \textbf{53.1} & \textbf{65.3} \\
        \midrule
        \multicolumn{8}{c}{\it{Object365v2 Dataset}} \\
        \midrule
        \multicolumn{2}{l|}{T: RetinaNet-R101} & 19.5 & 28.0 & 20.9 & 8.7 & 19.6 & 26.2 \\
        \multicolumn{2}{l|}{S: RetinaNet-R50}  & 16.7 & 24.4 & 17.8 & 7.2 & 16.7 & 22.0 \\
        \midrule
        \rowcolor{gray!20}
        \multicolumn{2}{l|}{CanKD}             & \textcolor{blue}{\bf{18.4(1.7$\uparrow$)}} & 26.6 & 19.7 & 7.7 & 18.3 & 24.9 \\
        \bottomrule
        \end{tabular}
    }
\end{table}


\section{Analysis}
\subsection{Ablation study of Can block and instance normalization}
To verify the impact of Can block and instance normalization, we compared CanKD against a variant that removes the Can block and another variant that removes the Can block and instance normalization. Our results are recorded in Table \ref{8}. We use RepPoints-X101 as the teacher model and RepPoints-R50 as the student model. To maintain consistency with the original model, the weight for all methods is set to five. Our results successfully demonstrate the effectiveness of both the Can block and instance normalization. Compared with a distillation method that uses only L2 loss, jointly employing instance normalization and the Can block improves the AP by 1.0 points.

\begin{table}[tb]
    \centering
    \caption{\textbf{Ablation study of instance normalization and the Can block.} We use RepPoints-X101 \cite{yang2019reppoints} as the teacher and RepPoints-R50 as the student.}
    \label{8}
    \resizebox{\linewidth}{!}{
        \begin{tabular}{lccccccc}
        \toprule
        \multicolumn{2}{l}{Method} & $AP$ & $AP_{50}$ & $AP_{75}$ & $AP_S$ & $AP_M$ & $AP_L$ \\
        \midrule
        \midrule
        \multicolumn{2}{l|}{w/o KD (baseline)}  & 38.6 & 59.6 & 41.6 & 22.5 & 42.2 & 50.4 \\
        \multicolumn{2}{l|}{L2 only} & 41.4 & 62.0 & 44.5 & 23.3 & 45.2 & 54.5 \\
        \multicolumn{2}{l|}{Instance Norm + L2} & 41.7 & 62.2 & 45.0 & 23.6 & 46.1 & 55.6 \\
        \rowcolor{gray!20}
        \multicolumn{2}{l|}{\textbf{CAN + Instance Norm + L2(CanKD)}} & 42.4 & 62.9 & 45.6 & 24.1 & 46.5 & 56.4 \\
        \bottomrule
        \end{tabular}
    }
\end{table}

\begin{table}[tb]
    \centering
    \caption{\textbf{Ablation study of different affinity functions.} We use RepPoints-X101 \cite{yang2019reppoints} as the teacher and RepPoints-R50 as the student to test the sensitivity of the affinity function.}
    \label{4}
    \resizebox{\linewidth}{!}{
        \begin{tabular}{lccccccccc}
        \toprule
        \multicolumn{2}{l}{Method} & $Softmax$ & $\theta,\phi$ & $AP$ & $AP_{50}$ & $AP_{75}$ & $AP_S$ & $AP_M$ & $AP_L$ \\
        \midrule
        \midrule
        \multicolumn{2}{l|}{Gaussian} & $\checkmark$ & $\times$ & 41.8 & 62.2 & 45.0 & 24.1 & 46.0 & 55.9 \\
        \multicolumn{2}{l|}{Embedded Gaussian} & $\checkmark$ & $\checkmark$ & 41.9 & 62.3 & 45.3 & 24.4 & 46.4 & 55.3 \\
        \rowcolor{gray!20}
        \multicolumn{2}{l|}{Dot product} & $\times$ & $\checkmark$ & 42.4 & 62.9 & 45.6 & 24.1 & 46.5 & 56.4 \\
        \bottomrule
        \end{tabular}
    }
    \vspace{-5mm}
\end{table}

\begin{figure*}[tb]
    \centering
    \includegraphics[width=\linewidth]{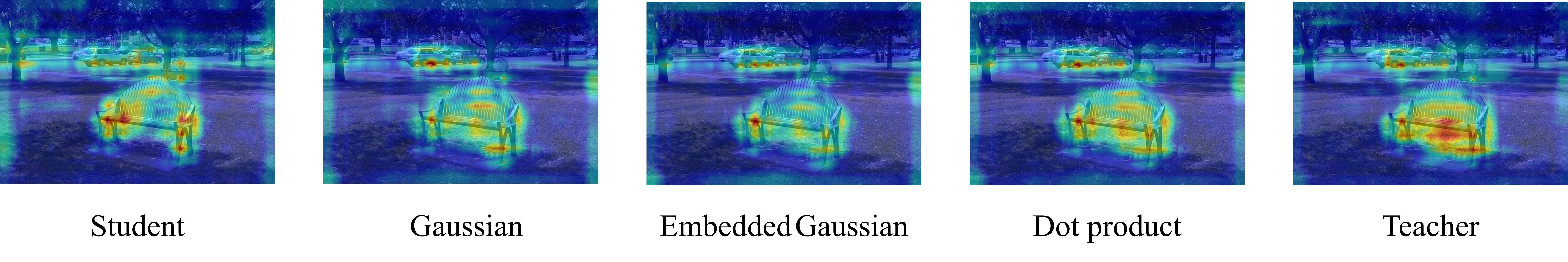}
    \caption{\textbf{Visualization heatmaps for different affinity functions.} We use RepPoints-X101 \cite{yang2019reppoints} as the teacher and RepPoints-R50 as the student. These heatmaps are generated from P6 in the FPN layers}
    \label{fig.1}
    \vspace{-3mm}
\end{figure*}

\subsection{Sensitivity study of the affinity function}
We replace the dot product affinity function in equation \eqref{eq:relation_function} with two alternative functions: the Gaussian function \cite{buades2005non,tomasi1998bilateral} and the embedded Gaussian function \cite{vaswani2017attention}.\footnote{The details of these functions are available in Section 2 in the supplementary.} We use RepPoints-X101 as the teacher model and RepPoints-R50 as the student model. As shown in Table \ref{4}, the dot product approach shows a clear advantage over the other two Gaussian methods. We conjecture that this superiority may stem from the dot product's positive influence on cross-attention. 

In Figure \ref{fig.1}, we present the heatmaps for each method from intermediate layers. These visualizations indicate that, compared with the dot product approach, the other two Gaussian methods are less effective in capturing the high-activation regions (the warmer-colored areas in the heatmaps) in the teacher's feature map. We conjecture that introducing the \textit{Softmax} layer in the first two methods normalizes the attention scores obtained from the affinity function into the range [0,1], thereby reducing the difference in attention values across pixels. Meanwhile, the dot product approach computes attention more directly, facilitating backpropagation. Consequently, we believe these factors account for the superior performance of the dot product strategy within our distillation framework.

\subsection{Sensitivity study of weight hyperparameter}
In equation \eqref{eq:total}, we use a hyperparameter $\mu$ to balance the magnitudes of the distillation loss and the task loss in the student model. Here, we explore alternative parameter values to investigate the impact of different hyperparameter settings on the model's performance.


As shown in Table \ref{5}, in the object detection task with this teacher-student pairing, setting $\mu=5$ yields the highest performance, improving the AP by 3.8 points over the baseline. In contrast, $\mu=10$ gives the lowest performance with an improvement of only 3.0 points. As shown in Table \ref{6}, $\mu=10$ achieves the best performance for the semantic segmentation task, improving mIoU by 0.64 points. Overall, the semantic segmentation task model exhibits less sensitivity to this parameter. Furthermore, according to the log data, when the value of the distillation loss is close to that of the task loss, our model easily achieves peak performance.

\begin{table}[tb]
    \centering
    \caption{\textbf{Ablation study of different hyperparameters for the object detection task.} We use RepPoints-X101 \cite{yang2019reppoints} as the teacher and RepPoints-R50 as the student.}
    \label{5}
    \resizebox{0.8\linewidth}{!}{
        \begin{tabular}{lccccccc}
        \toprule
        \multicolumn{2}{l}{Weight} & $AP$ & $AP_{50}$ & $AP_{75}$ & $AP_S$ & $AP_M$ & $AP_L$ \\
        \midrule
        \midrule
        \multicolumn{2}{l|}{Baseline}  & 38.6 & 59.6 & 41.6 & 22.5 & 42.2 & 50.4 \\
        \midrule
        \multicolumn{2}{l|}{$\mu=2$}  & 42.3 & 63.0 & 45.5 & 24.1 & 46.5 & 56.2 \\
        \rowcolor{gray!20}
        \multicolumn{2}{l|}{$\mu=5$} & 42.4 & 62.9 & 45.6 & 24.1 & 46.5 & 56.4 \\
        \multicolumn{2}{l|}{$\mu=8$} & 41.8 & 62.0 & 45.2 & 24.3 & 46.4 & 54.5 \\
        \multicolumn{2}{l|}{$\mu=10$} & 41.6 & 61.7 & 45.0 & 23.3 & 46.1 & 55.5 \\
        \bottomrule
        \end{tabular}
    }
\end{table}

\begin{table}[tb]
    \centering
    \caption{\textbf{Ablation study of different hyperparameters for the semantic segmentation task.} We use PSPNet-R101 \cite{zhao2017pyramid} as the teacher and DeepLabV3-R18 \cite{chen2018encoder} as the student.}
    \label{6}
        \begin{tabular}{lcccc}
        \toprule
        \multicolumn{2}{l}{Weight} & mIoU(\%) & aAcc(\%) & mAcc(\%) \\
        \midrule
        \midrule
        \multicolumn{2}{l|}{Baseline} & 76.70 & 95.89 & 84.29 \\
        \midrule
        \multicolumn{2}{l|}{$\mu=2$} & 76.96 & 96.33 & 85.03 \\
        \multicolumn{2}{l|}{$\mu=5$} & 77.28 & 96.02 & 85.08 \\
        \multicolumn{2}{l|}{$\mu=8$} & 77.00 & 96.03 & 84.97 \\
        \rowcolor{gray!20}
        \multicolumn{2}{l|}{$\mu=10$} & 77.34 & 96.04 & 85.49 \\
        \bottomrule
        \end{tabular}
        \vspace{-3mm}
\end{table}

\begin{figure}[tb]
    \centering
    \includegraphics[width=\linewidth]{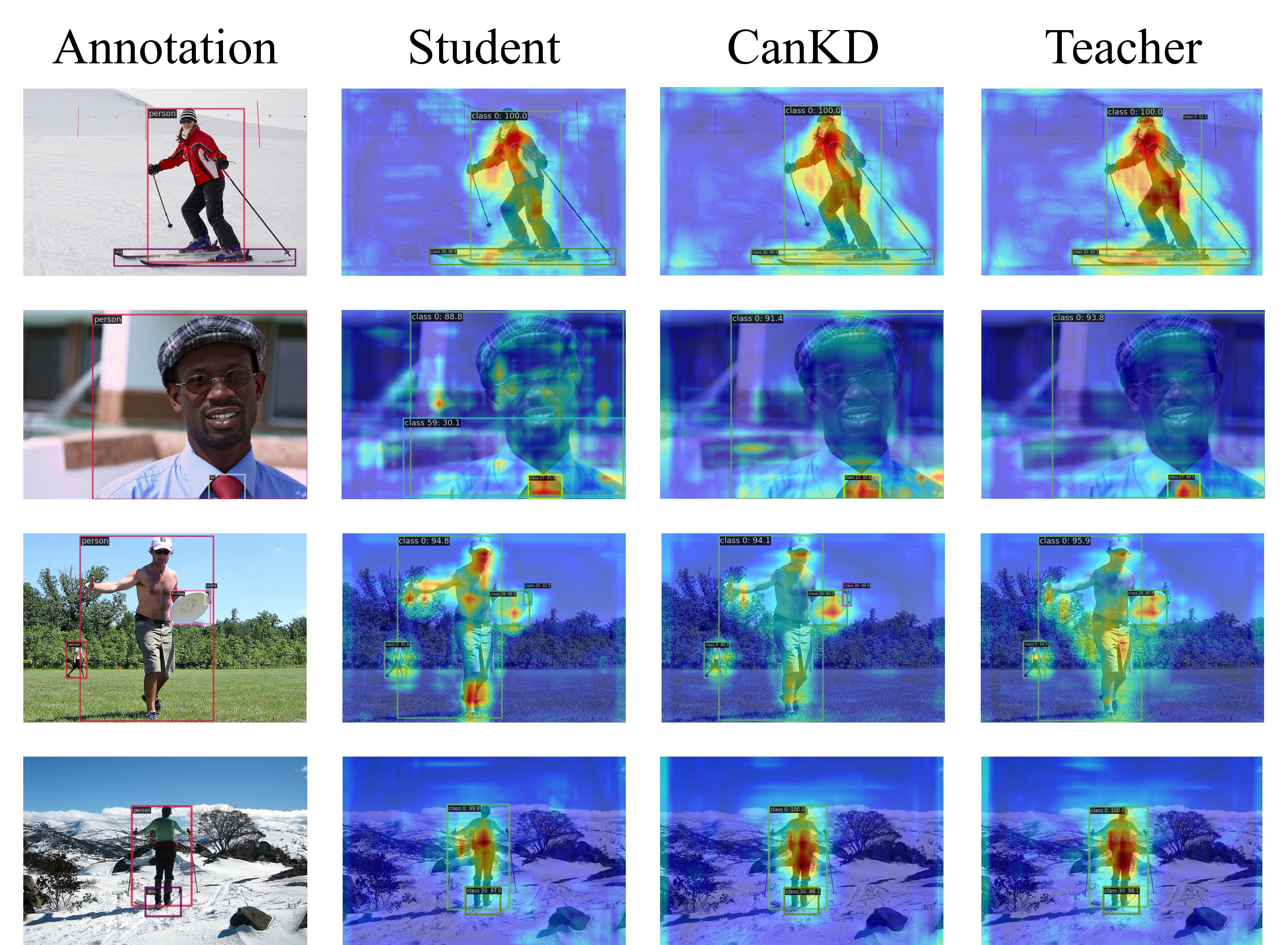}
    \caption{\textbf{Visualization of student heatmaps on P6 in FPN layers, where the student is distilled with CanKD heatmaps and teacher heatmaps for the COCO validation dataset.} These heatmaps are selected from T:FasterRCNN-R101-S:FasterRCNN-R50 and T:RepPoints-X101-S:RepPoints-R50.}
    \label{Fig.3}
    \vspace{-3mm}
\end{figure}

\begin{figure}[tb]
    \centering
    \includegraphics[width=\columnwidth]{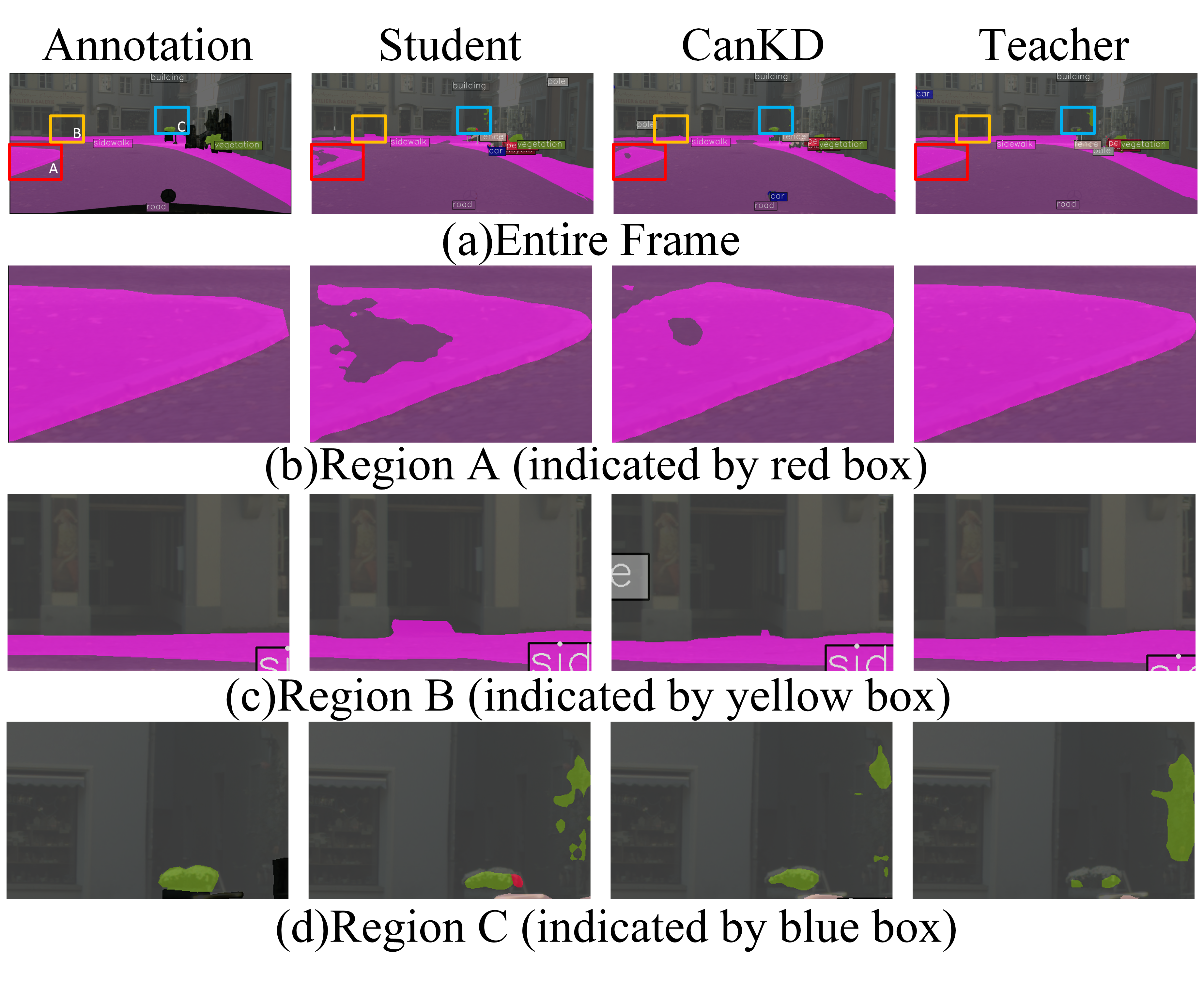}
    \caption{\textbf{Visualization of prediction results for semantic segmentation task in the Cityscapes validation dataset.} The teacher is PSPNet-R101 and the student is PSPNet-R50.}
    \label{Fig.4}
    \vspace{-3mm}
\end{figure}

\subsection{Visualization}
In Figures \ref{Fig.3} and \ref{Fig.4}, we show the heatmaps and prediction maps of CanKD for object detection and semantic segmentation. The student model distilled with CanKD effectively inherits the teacher’s knowledge for dense prediction tasks. Compared with the original student feature maps, the distilled model’s activations better align with those of the teacher. Moreover, when the student produces correct outputs but the teacher errs, our distilled model preserves the student’s accurate predictions.\footnote{Additional visualization examples are provided in Section 8 of the supplementary.} We attribute this to the cross-attention mechanism in the Can module, which builds robust latent relations between student and teacher pixels—unlike conventional attention methods that enforce pixel-level, one-to-one correspondences.

\section{Conclusion}
In this work, we proposed Cross-Attention-based Non-local Knowledge Distillation (CanKD). By leveraging cross-attention mechanisms, this novel distillation method enhances knowledge transfer in teacher-to-student models. CanKD allows each pixel in the student feature map to dynamically consider all pixels in the teacher feature map. This approach captures complex pixel-level relationships, leading to more effective feature learning and improved model performance.

Through extensive experiments on dense prediction tasks, we demonstrated that CanKD significantly outperforms state-of-the-art feature distillation methods and hybrid distillation approaches. Our method achieves superior performance while introducing only an additional loss function, rendering it computationally efficient compared with existing attention-based distillation techniques.
The results highlight CanKD's potential as a new paradigm for attention-guided knowledge distillation. 

Finally, we discuss the limitations of our proposed method. Attention-based distillation methods share a common drawback: they are highly sensitive to pixel-level variations. Consequently, the widespread use of bilinear interpolation for H×W alignment can degrade their performance. Although CanKD performs slightly worse than complex multi-loss distillation methods on large models, its modular design makes it a promising component for integration into such frameworks, potentially enhancing their overall distillation effectiveness.\footnote{A more detailed discussion of the limitations is provided in Section 11 of the supplementary material.}


{
    \small
    \bibliographystyle{ieeenat_fullname}
    \bibliography{CanKD}
}

\end{document}


\maketitle
\section{Details about training strategy}
In this paper, we verify the superior performance of CanKD by conducting experiments with various model architectures on both object detection and semantic segmentation tasks. We train all models on NVIDIA RTX 6000 Ada 48GB with 2 GPUs. In the object detection task, following the guidelines provided by the mmdetection\cite{mmdetection} documentation and adhering to the linear scaling rule\cite{goyal2018accuratelargeminibatchsgd}, we set the learning rate to 0.005. Meanwhile, following the mmdetection $2\times$ training strategy, we employ a multi-step learning rate decay at epochs 16 and 22 with a decay factor of 0.1. In the semantic segmentation task, following the mmsegmentation\cite{mmseg2020} official training strategy of an 80K schedule, we set the learning rate to 0.01 with a weight decay of 0.0005 and use a polynomial function to decay our learning rate. The number of parameters and flops in the segmentation task results are all calculated by \textit{thop}.

\section{Details about affinity function}
In this section, we provide a detailed analysis of two alternative affinity functions that deviate from our original method: the Gaussian method\cite{buades2005non,tomasi1998bilateral} and the Embedding Gaussian method\cite{vaswani2017attention}.

\subsection{Gaussian affinity function}
From the perspective of early non-local methods, directly processing two feature maps with a Gaussian function represented a straightforward choice. We omit two $1\times1$ modules from $\theta$ and $\phi$, and directly compute the affinity between the student's feature map and the teacher's feature map. The Gaussian affinity function is determined by:
\begin{equation}
    \label{1}
    \xi(\bm{x}_i, \bm{y}_j) = \exp(\bm{x}_i^\top\bm{y}_j).
\end{equation}
where, $\bm{x}_i$ and $\bm{y}_j$ represent the feature vectors at the position $i$ in the student feature map and the position $j$ in the teacher feature map, respectively. The normalization factor for the Gaussian affinity function is \(C = \sum_{j}{\xi(\bm{x}_i, \bm{y}_j})\). 
In terms of implementation, this operation can be easily carried out by applying a \textit{Softmax} layer.

\subsection{Embedded Gaussian affinity function}
Analogous to self-attention in Transformers, the Embedded Gaussian method introduces an embedding space. Specifically, the student and teacher feature maps are projected into the embedding space via two $ 1\times 1$ convolutional layers, denoted as $\theta$ and $\phi$. The affinity is then calculated using a Gaussian function, as shown in equation \eqref{2equation}.
\begin{equation}
    \xi(\bm{x}_i, \bm{y}_j) = \exp\left(\theta(\bm{x}_i)^{\top}\phi(\bm{y}_j)\right).
    \label{2equation}
\end{equation}
The normalization factor for the embedded Gaussian affinity function is also \(C = \sum_{j}{\xi(\bm{x}_i, \bm{y}_j})\). 

\section{Natural corrupted augmentation analysis}
Following the \cite{michaelis2019winter}, we evaluate the student RetinaNet-R50 detector, which CanKD trained in the COCO-C dataset. The COCO-C dataset is derived from the COCO validation dataset by applying four types of corruption, i.e., transformations—noise, blurring, weather, and digital corruption, to evaluate the model’s robustness. Each corruption category consists of multiple corruption methods, each with six levels of severity. The test results are summarized in Table \ref{7}. The results show that CanKD exhibits substantially higher robustness than the other methods. Specifically, it achieves a 2.2 improvement in mPC and a 2.7 improvement in rPC compared to the benchmark.

\begin{table}[tbp]
    \centering
    \caption{\textbf{Result of robust object detection via CanKD on COCO-C dataset.}}
    \label{7}
        \begin{tabular}{c|ccc}
        \toprule
        Method & $AP_{\texttt{clean}}$ & mPC & rPC \\
        \midrule
        \midrule
        RetinaNet-R50 & 37.4 & 18.3 & 48.9  \\
        \midrule
        FGD & 39.6 & 20.3 & 51.3  \\
        DiffKD & 39.7 & 20.3 & 51.1 \\
        \rowcolor{gray!20}
        CanKD & \textbf{39.8} & \textbf{20.5} & \textbf{51.6} \\
        \bottomrule
        \end{tabular}
\end{table}

\section{Analysis about maxpooling layer}
In this section, we examine whether different scales of the teacher feature maps after max pooling influence the performance of the CanKD. We experimented with max pooling at $4\times4$ and $8\times8$ scales. It is important to note that while reducing the pooling scale can facilitate the student model's more comprehensive acquisition of the teacher feature maps, it simultaneously increases the memory footprint on the hardware. The experimental results are presented in Table.\ref{2}.

\begin{table}[tbp]
    \centering
    \caption{\textbf{Ablation study on maxpooling scales.} We use RepPoints-X101\cite{yang2019reppoints} as teacher and RepPoints-R50 as student.}
    \label{2}
    \resizebox{\linewidth}{!}{
        \begin{tabular}{lccccccc}
        \toprule
        \multicolumn{2}{l}{Scaled} & $AP$ & $AP_{50}$ & $AP_{75}$ & $AP_S$ & $AP_M$ & $AP_L$ \\
        \midrule
        \midrule
        \multicolumn{2}{l|}{$2\times2$}  & \textbf{42.4} & \textbf{62.9} & \textbf{45.6} & \textbf{24.1} & 46.5 & \textbf{56.4} \\
        \multicolumn{2}{l|}{$4\times4$} & 42.0 & 62.3 & 45.5 & 24.0 & 46.2 & 55.0 \\
        \multicolumn{2}{l|}{$8\times8$} & 42.1 & 62.3 & 45.6 & 23.9 & \textbf{46.7} & 55.7 \\
        \bottomrule
        \end{tabular}
    }
\end{table}

From the results, we observe that employing a large-scale max pooling operation eliminates certain crucial information within the teacher feature maps. This reduction hampers the student feature maps from fully assimilating the teacher’s knowledge. Meanwhile, we suggest that employing a max pooling layer to select critical information from the teacher feature maps may not be the optimal choice. A more refined strategy could better preserve essential information while reducing the spatial resolution of the teacher feature maps.

\section{Analysis about residual connection}
In this section, we examine the role of the residual connection within the Can module, which is in function 5 in the original paper. We believe that directly comparing the feature map produced by the attention component of the Can module with the teacher’s original feature map would cause issues because we do not apply any additional operations to the teacher's feature map. Therefore, we retain the residual connection so that the student’s original feature map remains involved. The ablation study is in table.\ref{rc}.

\begin{table}[tbp]
    \centering
    \caption{\textbf{Ablation study on residual connection.} We use RepPoints-X101\cite{yang2019reppoints} as teacher and RepPoints-R50 as student.}
    \label{rc}
    \resizebox{\linewidth}{!}{
        \begin{tabular}{lccccccc}
        \toprule
        \multicolumn{2}{l}{Scaled} & $AP$ & $AP_{50}$ & $AP_{75}$ & $AP_S$ & $AP_M$ & $AP_L$ \\
        \midrule
        \midrule
        \multicolumn{2}{l|}{W/O residual connection} & 41.4 & 61.8 & 44.6 & 24.4 & 45.6 & 54.3 \\
        \multicolumn{2}{l|}{W/ residual connection}  & \textbf{42.4} & \textbf{62.9} & \textbf{45.6} & \textbf{24.1} & \textbf{46.5} & \textbf{56.4} \\
        \bottomrule
        \end{tabular}
    }
\end{table}

According to the experimental results, CanKD with a residual connection outperforms CanKD without a residual connection by a significant margin. Therefore, we conclude that the residual connection is indispensable in CanKD.


\section{Confusion matrix}
\begin{figure*}[tb]
    \centering
    \includegraphics[width=\linewidth]{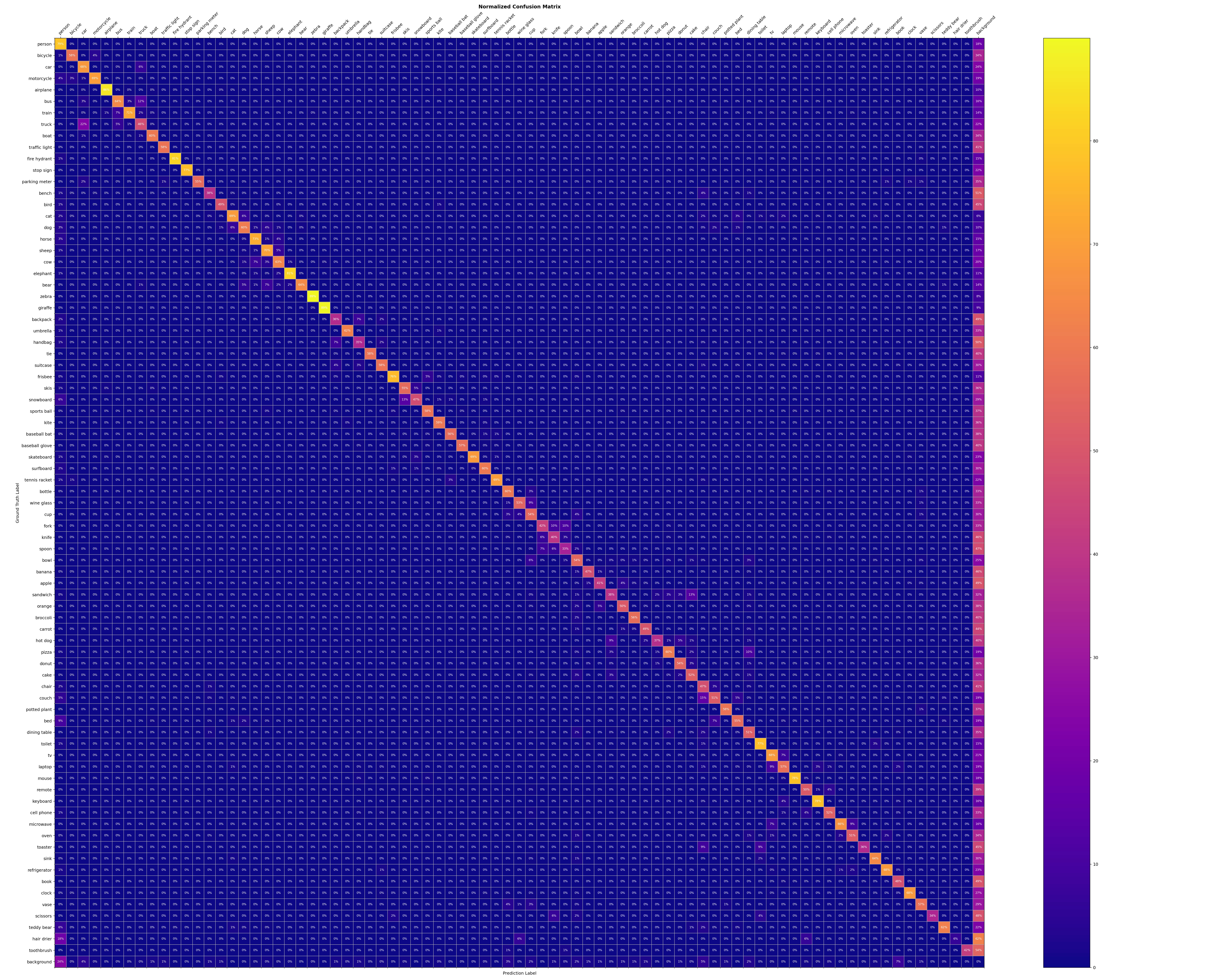}
    \caption{\textbf{Confusion matrix from FasterRCNN-R50.}}
    \label{fig.1}
\end{figure*}

\begin{figure*}[tb]
    \centering
    \includegraphics[width=\linewidth]{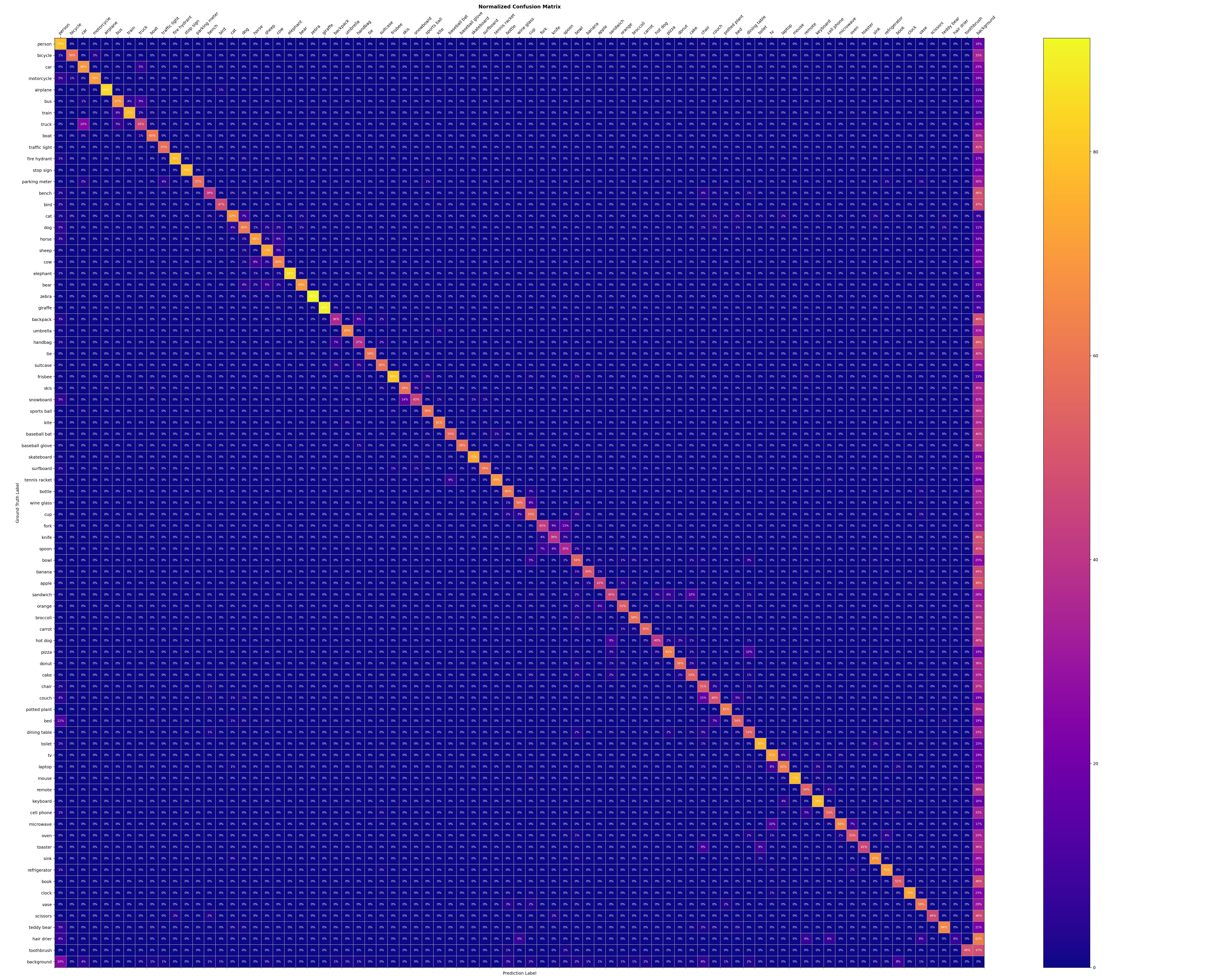}
    \caption{\textbf{Confusion matrix from FasterRCNN-R50 distilled with CanKD.}}
    \label{fig.2}
\end{figure*}

Here, we present the confusion matrices for the teacher models, the student models, and our proposed CanKD method. Here, Figure \ref{fig.1} is from the FasterRCNN-R50 \cite{ren2015faster} student model, and Figure \ref{fig.2} is from the FasterRCNN-R50 model distilled with CanKD.

\section{Detail for training}
\begin{figure*}[tb]
    \centering
    \includegraphics[width=\linewidth]{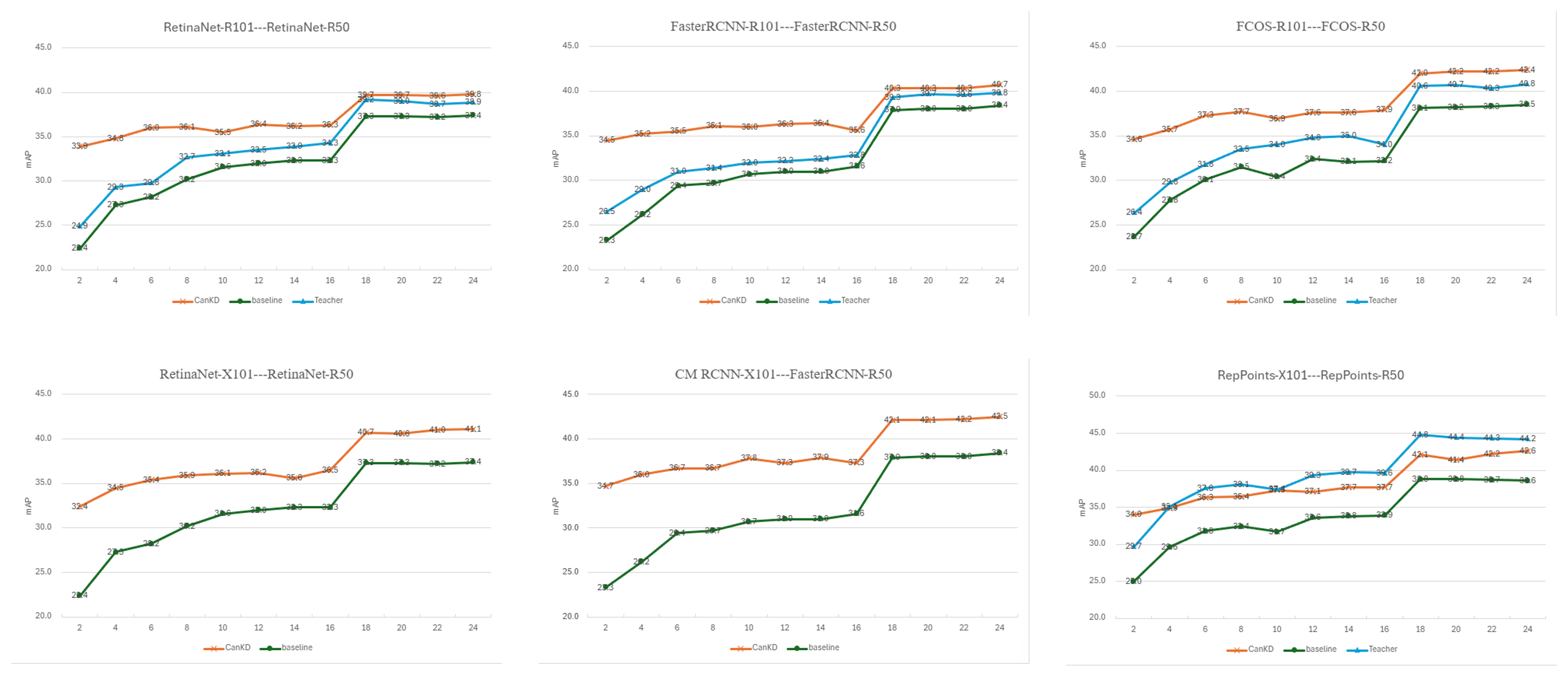}
    \caption{\textbf{Line chart about all student models, distilled student models, and teacher models.}}
    \label{fig.3}
\end{figure*}

By analyzing the output logs from each model, we generated the changes in mAP for each model throughout the training process. The figure is shown in Figure.\ref{fig.3}.

\begin{figure*}[tb]
    \centering
    \includegraphics[width=\linewidth]{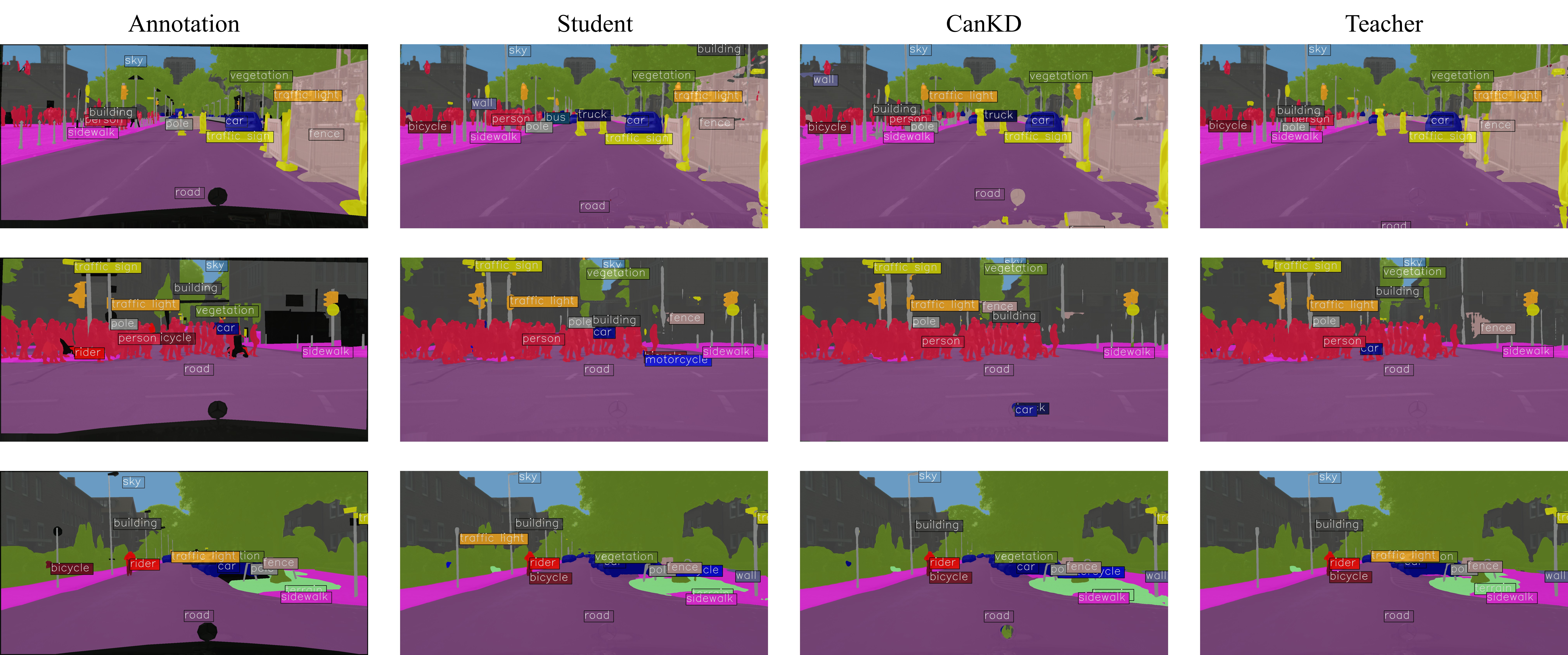}
    \caption{\textbf{Additional sampling from PSPNet-R18\cite{zhao2017pyramid}, ditilled PSPNet-R18 with CanKD and PSPNet-R101.} All of these figures are selected from Cityscapes val dataset.}
    \label{fig.4}
\end{figure*}

\begin{figure*}[tb]
    \centering
    \includegraphics[width=\linewidth]{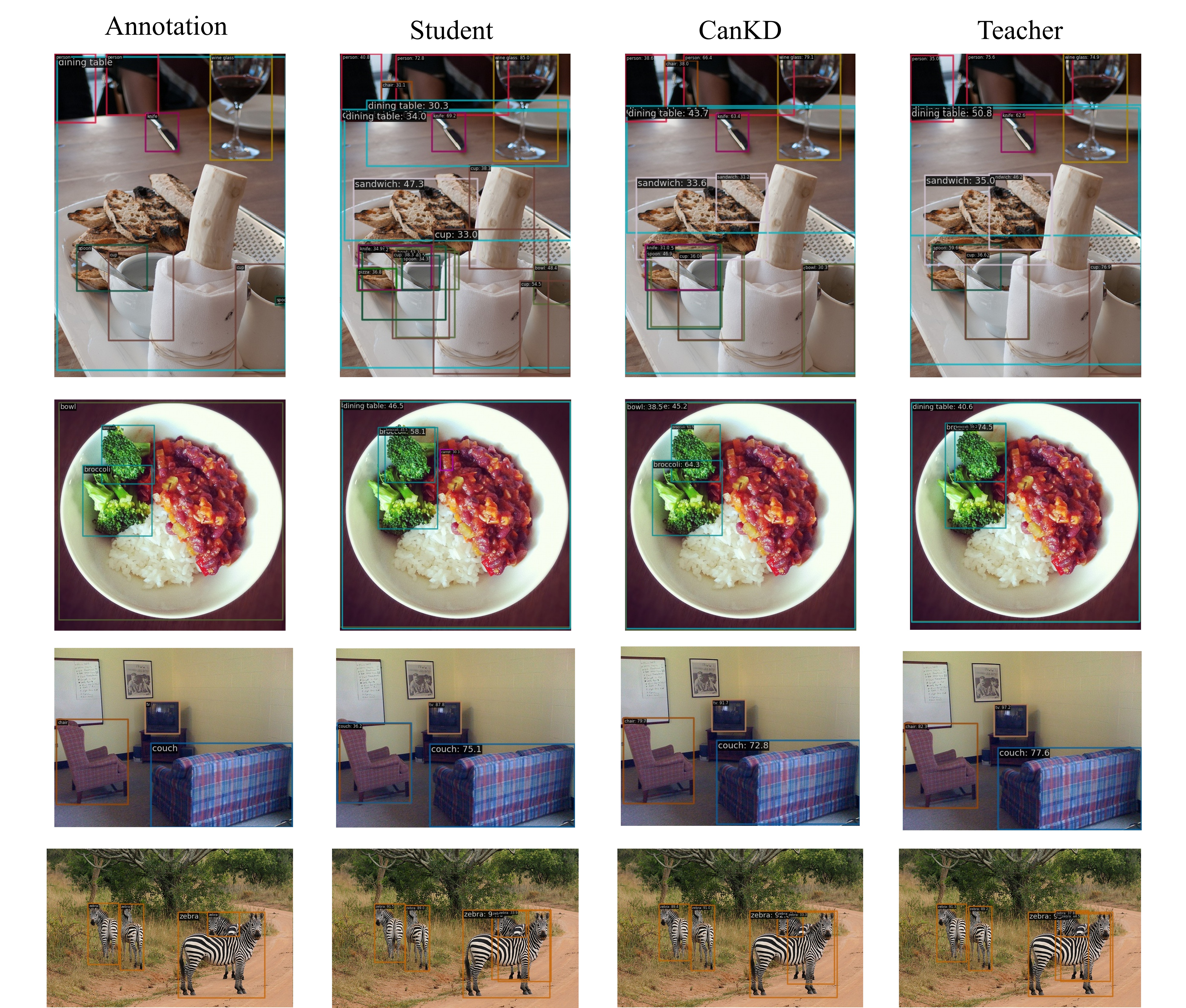}
    \caption{\textbf{Additional sampling from RepPoints-R50, ditilled RepPoints-R50 with CanKD and RepPoints-X101.} All of these figures are selected from COCO val 2017 dataset.}
    \label{fig.5}
\end{figure*}

\section{Visualization}
\begin{figure*}
    \centering
    \includegraphics[width=\linewidth]{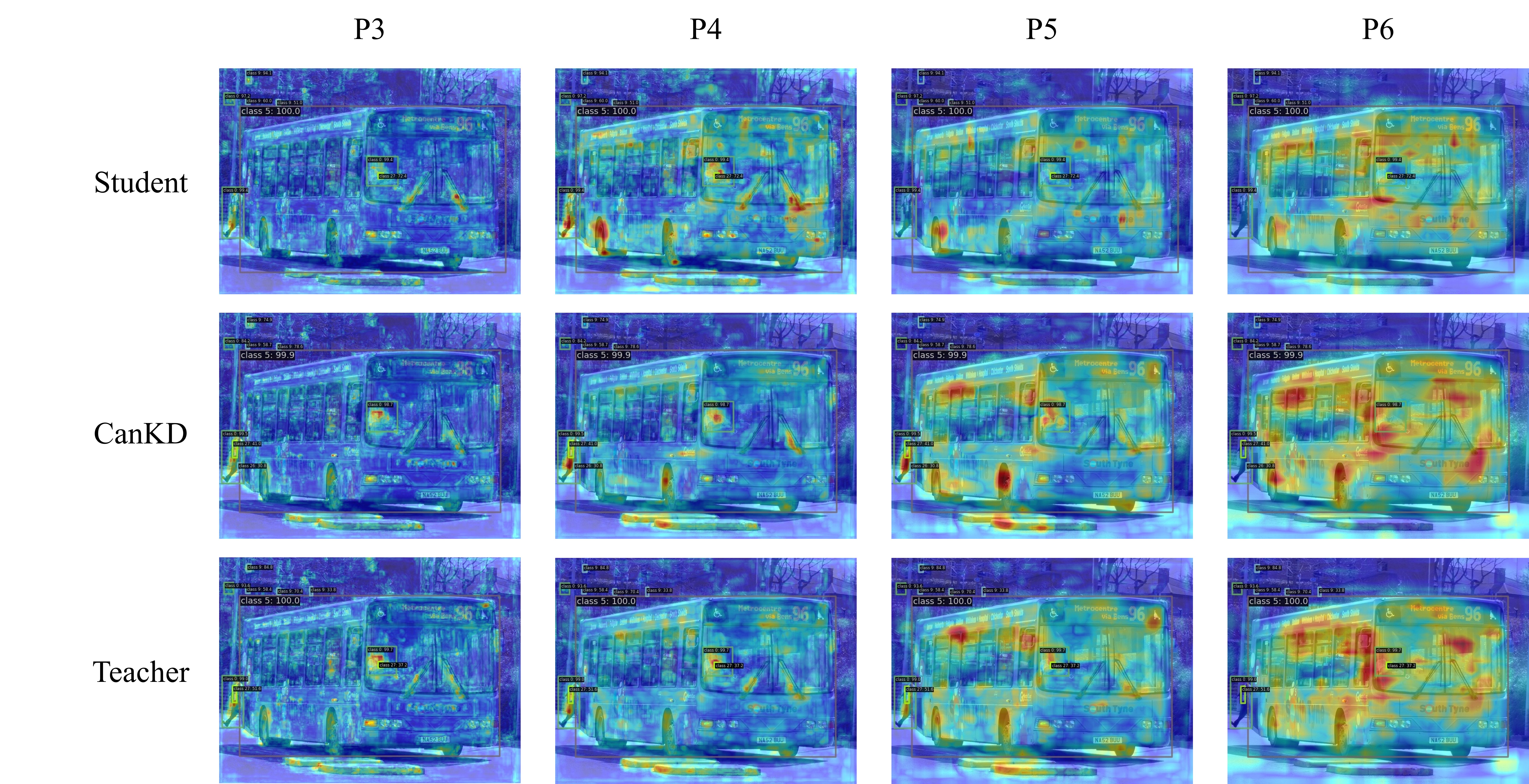}
    \caption{\textbf{Heatmaps from FasterRCNN-R50, ditilled FasterRCNN-R50 with CanKD and FasterRCNN-R101.} These figures are generated from P3 to P5 in FPN layers}
    \label{fig.6}
\end{figure*}

In Figure \ref{fig.5}, we present a series of visualization figures generated from the student model, CanKD, and the teacher model on the COCO val 2017 dataset\cite{lin2014microsoft} to demonstrate the performance and effectiveness of our method in object detection. Meanwhile, in Figure \ref{fig.4}, we also present a series of visualization figures generated from the student model, CanKD, and the teacher model on the Cityscapes validation dataset\cite{cordts2016cityscapes}. In Figure \ref{fig.6}, we present the heatmaps from different FPN layers for the student model, the distilled student model, and the teacher model. Compared to the original student model, these examples demonstrate that our method has better performance and effectiveness. 

\section{Experiments on classification task.}
\begin{figure}[h]
    \centering
    \includegraphics[width=\linewidth]{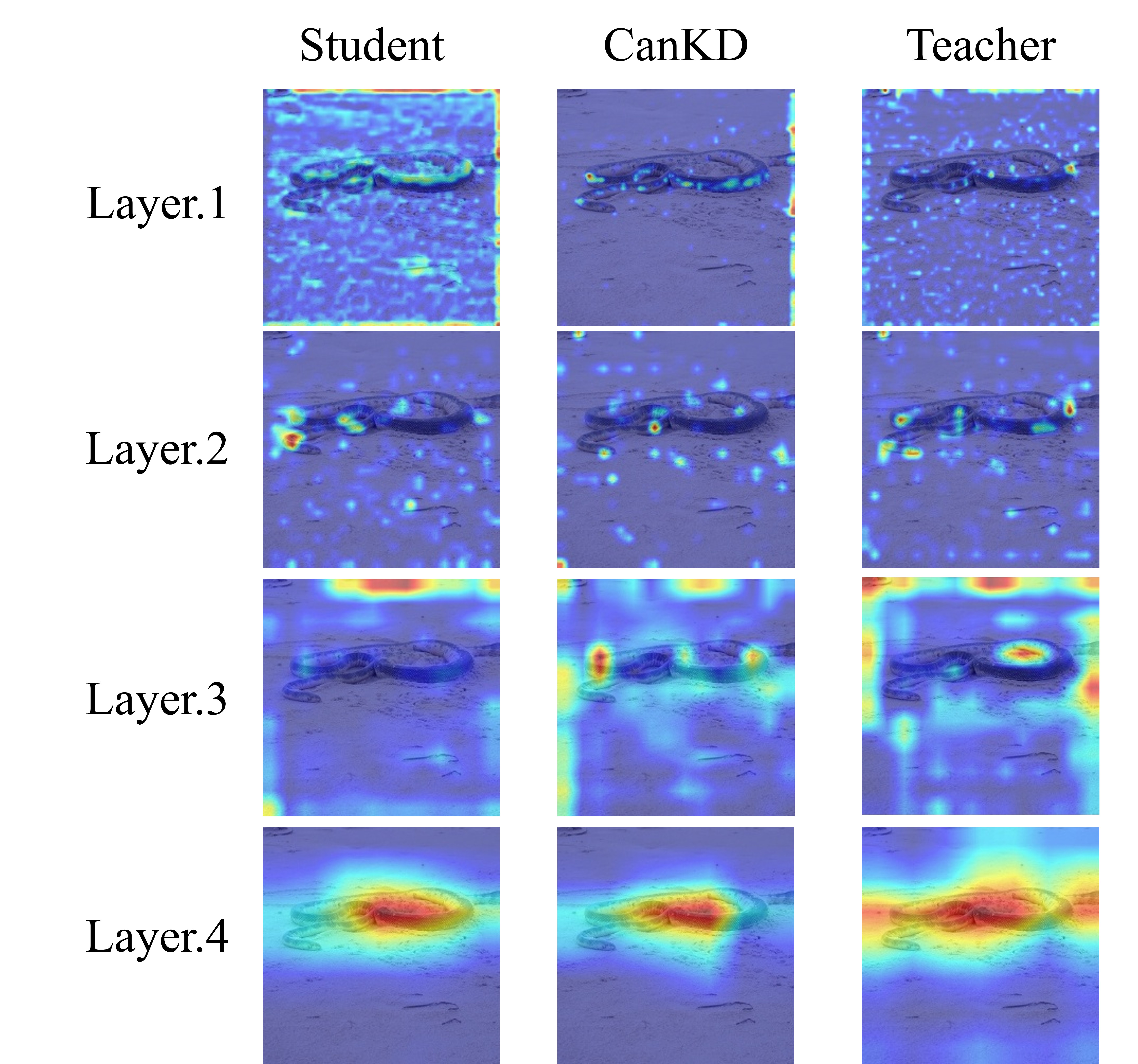}
    \caption{\textbf{Heatmaps from ResNet-18, ditilled ResNet-18 with CanKD and ResNet-32.} These figures are generated from Layer.1 to Layer.4 in backbone.}
    \label{cls_img}
\end{figure}

To complement the evaluation of CanKD on fundamental tasks, following the official schedule, we conducted experiments on the ImageNet-1K \cite{deng2009imagenet} dataset with two teacher–student model pairs, ResNet-34$\rightarrow$ResNet-18, ResNet-50$\rightarrow$MobileNetv1 in 100e. We choose the last layer output in backbone as our distillation position. The results are shown in table.\ref{cls}. The balanced weight $\mu$ in CanKD are all set to 5. However, CanKD is specifically designed for dense prediction tasks, where pixel-level feature alignment is critical. In contrast, classification tasks do not involve pixel-level alignment; therefore, the use of cross-attention may be unnecessary compared to traditional logit distillation methods that already achieve strong results. Meanwhile, CanKD outperforms w/o student, KD \cite{hinton2015distilling}, and AT \cite{zagoruyko2016paying}. This demonstrates that the performance gains of CanKD in dense prediction do not come at the expense of classification performance. The classification heat map images are shown in Figure \ref{cls_img} which generated by \textit{Grad-CAM} \cite{selvaraju2017grad}.

\begin{table}[tbp]
    \centering
    \caption{\textbf{Classification task on ImageNet-1K.} We use ResNet-34, ResNet-50 \cite{he2016deep} as teachers and Resnet-18, MobileNet-v1 \cite{howard2017mobilenets} students. All baselines results are from \cite{yang2022masked}.}
    \label{cls}
    \resizebox{\linewidth}{!}{
        \begin{tabular}{cccc|cccc}
        \toprule
        \multicolumn{2}{l|}{Method} & $Top1$ & $Top5$ & \multicolumn{2}{l|}{Method}  & $Top1$ & $Top5$ \\
        \midrule
        \multicolumn{2}{l|}{T:ResNet-34} & 73.62 & 91.59 & \multicolumn{2}{l|}{T:ResNet-50} & 76.55 & 93.06 \\
        \multicolumn{2}{l|}{T:ResNet-18} & 69.90 & 89.43 & \multicolumn{2}{l|}{S:MobileNetv1} & 69.21 & 89.02 \\
        \midrule
        \midrule
        \multicolumn{2}{l|}{KD \cite{hinton2015distilling}}  & 70.68 & \textbf{90.16} & \multicolumn{2}{l|}{KD \cite{hinton2015distilling}} & 70.68 & \textbf{90.30} \\
        \multicolumn{2}{l|}{AT \cite{zagoruyko2016paying}}  & 70.59 & 89.73 & \multicolumn{2}{l|}{AT \cite{zagoruyko2016paying}} & 70.72 & 90.03 \\
        \midrule
        \multicolumn{2}{l|}{CanKD} & \textbf{70.73} & 89.81 & \multicolumn{2}{l|}{CanKD} & \textbf{70.89} & 89.90 \\
        \bottomrule
        \end{tabular}
    }
\end{table}

\section{Analysis about parameter and computational complexity}
In this section, we analyze the parameter count and computational complexity of CanKD. We compare its parameters and FLOPs with several strong feature distillation methods in table.\ref{param}, and our results show that CanKD has fewer parameters and lower complexity than existing attention-based and mask-based distillation methods. Moreover, since CanKD is applied only during training, it does not affect the parameter count or complexity of the student model at inference time, further demonstrating that CanKD is a lightweight yet effective attention-based distillation approach.

\begin{table}[tbp]
    \centering
    \caption{\textbf{Analysis on parameter and FLOPs.} We use \textit{thop} library to calculate params and FLOPs in same student and teacher's input with 2$\times$256$\times$64$\times$64. Because FGD \cite{yang2022focal} including bounding box distillation, it's hard to clarify the complexity.}
    \label{param}
    \resizebox{\linewidth}{!}{
        \begin{tabular}{c|cc|cc}
        \toprule
        Method & \multicolumn{2}{l}{KD method} & FLOPs(G) & Params(M)  \\
        \midrule
        \midrule
        FKD \cite{zhang2020improve} & \multicolumn{2}{c}{Attention} & 3.23 & 0.46  \\
        MGD \cite{yang2022masked} & \multicolumn{2}{c}{Mask} & 9.66 & 1.18 \\ 
        FGD \cite{yang2022focal} & \multicolumn{2}{c}{Attention} & - & 0.14 \\ 
        \midrule
        CanKD & \multicolumn{2}{c}{Attention} & \textbf{1.09} & \textbf{0.13}  \\
        \bottomrule
        \end{tabular}
    }
\end{table}

\section{Limitation of CanKD}
Attention-based distillation methods share a common drawback: when the teacher and student feature maps differ in spatial resolution (H×W), performance degradation occurs. We demonstrate this phenomenon using a simple teacher–student pair in Table.\ref{limit}. Furthermore, from the distillation results of Dino-Swin-L and Dino-Swin-B, CanKD shows only marginal improvement for this pair, which previous research attribute to the substantial architectural gap between Transformer-based backbones, where a simple CNN-based approach cannot be effectively applied \cite{yang2024vitkd}. The observation that CWD \cite{shu2021channel} performs even worse than the baseline further corroborates this finding.
\begin{table}[tb]
    \centering
    \caption{The experiment result with mismatch H$\times$W}
    \label{limit}
    \resizebox{\linewidth}{!}{
        \begin{tabular}{lccccccc}
        \toprule
        \multicolumn{2}{l}{Method} & $AP$ & $AP_{50}$ & $AP_{75}$ & $AP_S$ & $AP_M$ & $AP_L$ \\
        \midrule
        \multicolumn{2}{l|}{T:MaskRCNN-SwinS}  & 48.2 & 69.8 & 52.8 & 32.1 & 51.8 & 62.7 \\
        \multicolumn{2}{l|}{S:Retina-R50}  & 37.4 & 56.7 & 39.6 & 20.0 & 40.7 & 49.7 \\
        \midrule
        \multicolumn{2}{l|}{PKD \cite{cao2022pkd}}  & 41.5 & 60.6 & 44.1 & 22.9 & 45.2 & 56.4 \\
        \multicolumn{2}{l|}{DetKDS \cite{lidetkds}}  & 41.4 & 60.8 & 44.4 & 23.4 & 45.1 & 55.5 \\
        \midrule
        \multicolumn{2}{l|}{CanKD}  & 40.5 & 59.0 & 43.3 & 22.5 & 44.7 & 54.2 \\
        \bottomrule
        \end{tabular}
    }
    \vspace{-5mm}
\end{table}

{
    \small
    \bibliographystyle{ieeenat_fullname}
    \bibliography{supplement}
}